\pdfoutput=1
\documentclass[11pt,letterpaper]{article}
\usepackage[ruled,vlined,commentsnumbered,titlenotnumbered]{algorithm2e}

\usepackage{url}
\usepackage{fullpage}

\usepackage[subtle,mathspacing=normal]{savetrees}
\usepackage{amsmath}
\usepackage{amssymb}
\usepackage{bbm}
\usepackage{amsthm}
\usepackage{color}
\usepackage{array}
\usepackage{xy}
\usepackage{setspace}
\usepackage{varwidth}%
\usepackage{multicol}
\usepackage{mdframed}
\usepackage{hyperref}
\usepackage{thm-restate}
\usepackage{todonotes}
\usepackage{tikz}
\usepackage{pgfplots}
\pgfplotsset{compat=1.18}
\usetikzlibrary{positioning,shapes.geometric,arrows.meta,calc,fit,backgrounds}
\usepackage{standalone}
\usepackage[capitalize]{cleveref}
\crefname{step}{Step}{Steps}
\Crefname{step}{Step}{Steps}
\crefname{thm}{Theorem}{Theorems}
\Crefname{thm}{Theorem}{Theorems}
\crefname{lem}{Lemma}{Lemmas}
\Crefname{lem}{Lemma}{Lemmas}
\crefname{prop}{Proposition}{Propositions}
\Crefname{prop}{Proposition}{Propositions}
\crefname{clm}{Claim}{Claims}
\Crefname{clm}{Claim}{Claims}
\crefname{cor}{Corollary}{Corollaries}
\Crefname{cor}{Corollary}{Corollaries}
\crefname{conj}{Conjecture}{Conjectures}
\Crefname{conj}{Conjecture}{Conjectures}
\crefname{rem}{Remark}{Remarks}
\Crefname{rem}{Remark}{Remarks}
\crefname{ex}{Example}{Examples}
\Crefname{ex}{Example}{Examples}
\crefname{prob}{Problem}{Problems}
\Crefname{prob}{Problem}{Problems}
\crefname{proto}{Protocol}{Protocols}
\Crefname{proto}{Protocol}{Protocols}
\crefname{claim}{Claim}{Claims}
\Crefname{claim}{Claim}{Claims}
\crefname{conjecture}{Conjecture}{Conjectures}
\Crefname{conjecture}{Conjecture}{Conjectures}
\crefname{observation}{Observation}{Observations}
\Crefname{observation}{Observation}{Observations}
\crefname{problem}{Problem}{Problems}
\Crefname{problem}{Problem}{Problems}
\crefname{protocol}{Protocol}{Protocols}
\Crefname{protocol}{Protocol}{Protocols}
\usepackage{threeparttable}

\newtheorem{thm}{Theorem}[section]

\theoremstyle{remark}

\theoremstyle{plain}
\newtheorem{theorem}[thm]{Theorem}
\newcommand{\DefinitionBodyFont}{\normalfont\itshape}
\newcommand{\DefinitionHeadingFont}{\normalfont\bfseries}
\newtheoremstyle{definitionstyle}
  {3pt}%
  {3pt}%
  {\DefinitionBodyFont}%
  {}%
  {\DefinitionHeadingFont}%
  {.}%
  {.5em}%
  {\thmname{#1}\thmnumber{ #2}\thmnote{ (#3)}}
\theoremstyle{definitionstyle}
\newtheorem{definition}[thm]{Definition}
\theoremstyle{remark}
\newtheorem{lemma}[thm]{Lemma}

\theoremstyle{remark}
\newtheorem{remark}[thm]{Remark}

\makeatletter
\newcommand{\UseDefinitionStyleForTheorems}{%
  \let\th@plain\th@definitionstyle
  \let\th@remark\th@definitionstyle
  \let\th@definition\th@definitionstyle
}
\makeatother
\newcommand{\defn}[1]{\textbf{\upshape\boldmath #1}}
\newcommand{\MathScriptStyle}{}
\let\OriginalMathPrime\prime
\newcommand{\UseScriptScriptStyle}{%
  \renewcommand{\MathScriptStyle}{\scriptscriptstyle}%
  \renewcommand{\prime}{{\scriptscriptstyle\OriginalMathPrime}}%
}

\ExplSyntaxOn
\NewDocumentCommand{\StyleRestatedTheorem}{m}{
  \tl_set:Nv \l_tmpa_tl {thmt@stored@#1}
  \regex_replace_all:nnN
    {\cU.}{\c{RestatedSuperscript}} \l_tmpa_tl
  \regex_replace_all:nnN
    {\cD.}{\c{RestatedSubscript}} \l_tmpa_tl
  \cs_gset:cpx {thmt@stored@#1}{\exp_not:V \l_tmpa_tl}
}
\ExplSyntaxOff
\renewcommand{\paragraph}[1]{\vspace{.2 cm} \noindent \textbf{#1}}

\usepackage{thm-restate}

\usetikzlibrary{matrix,arrows}
\usetikzlibrary{positioning}
\usetikzlibrary{decorations}
\usetikzlibrary{decorations.pathreplacing}
\usetikzlibrary{shapes.geometric, chains, calc, fit, matrix}
\tikzstyle{system}=[rectangle,draw,fill=lightgray,minimum height=0.8cm,minimum
            width=0.8cm,thick]
\tikzstyle{BC}=[system]
\tikzstyle{resource}=[system]
\tikzstyle{RO}=[resource, minimum width=1cm]
\tikzstyle{protocol}=[circle, inner sep=0.7mm, draw]
\tikzstyle{simulator}=[circle, inner sep=0.7mm, draw]
\tikzstyle{memory}=[resource]
\tikzstyle{distinguisher}=[resource,fill=white,minimum width=3.5cm,
minimum height=1.2cm]
\tikzstyle{link}=[]

\newcommand{\eps}{\varepsilon}

\usepackage{accents}

\usepackage[
    backend=bibtex,
    style=alphabetic,
    maxnames=10,
    maxalphanames=10,
]{biblatex}
\usepackage{complexity}

\newmdenv[
    skipabove=\smallskipamount,
    skipbelow=\smallskipamount,
    innertopmargin=6pt,
    innerbottommargin=6pt,
    innerleftmargin=8pt,
    innerrightmargin=8pt,
    linewidth=0.8pt,
    roundcorner=2pt,
]{protocolbox}

\newcommand{\bbN}{\mathbb{N}}

\newfunc{\RAM}{RAM}
\newfunc{\RAMDel}{RAMDel}
\newfunc{\RAMSNARG}{RAM~SNARG}
\newfunc{\SNARG}{SNARG}
\newfunc{\SNAP}{SNAP}
\newfunc{\seBARG}{seBARG}

\newfunc{\flSEH}{flSEH}
\newfunc{\HT}{HT}

\usepackage{microtype}
\usepackage{graphicx}
\usepackage{subcaption}
\usepackage{booktabs} %

\usepackage{hyperref}

\usepackage{mathtools}
\usepackage[olditem,oldenum]{paralist}
\usepackage{enumitem}
\usepackage{ragged2e}

\usepackage{placeins}

\newcommand{\cv}{\mathsf{CVAL}}
\newcommand{\CSATRelation}{\Relation_{\MathScriptStyle \mathsf{CSAT}}}

\newcommand{\pro}{\mathbf{P}}
\newcommand{\ver}{\mathbf{V}}

\newcommand{\CvalRelation}{\Relation_{\MathScriptStyle \cv}^{\MathScriptStyle \Oracle}}

\newcommand{\bits}{\{0,1\}}
\newcommand{\Language}{\mathcal{L}}
\newcommand{\Relation}{R}
\newcommand{\GetLanguage}[1]{\Language_{\MathScriptStyle #1}}
\newcommand{\Instance}{\mathbbm{x}}
\newcommand{\InstancePre}{\Instance_{\MathScriptStyle \mathrm{pre}}}
\newcommand{\InstancePost}{\Instance_{\MathScriptStyle \mathrm{post}}}
\newcommand{\CircuitInput}{x}

\newcommand{\InstanceSize}{n}
\newcommand{\Witness}{\mathbbm{w}}
\newcommand{\Circuit}{C}

\newcommand{\Oracle}{O}

\newcommand{\OracleCircuit}{\Circuit^{\MathScriptStyle \Oracle}}

\newcommand{\CircuitSize}{S}
\newcommand{\OutputSize}{m}
\newcommand{\OutputValue}{a}

\newcommand{\ComputedFunction}{f}

\newcommand{\OracleGateCount}{\mathsf{q}_{\scriptscriptstyle \Circuit}}

\newcommand{\QueryAnswerTrace}{\mathcal{Q}}

\newcommand{\OracleQuery}{\mathsf{qry}}
\newcommand{\OracleAnswer}{\mathsf{ans}}
\newcommand{\QueryInputs}[1]{#1_{\MathScriptStyle \mathrm{in}}}
\newcommand{\QueryAnswers}[1]{#1_{\MathScriptStyle \mathrm{out}}}
\newcommand{\BigO}[1]{O(#1)}

\newcommand{\SoftO}[1]{\widetilde{O}(#1)}

\newcommand{\OracleLanguage}{\Language^{\MathScriptStyle \Oracle}}
\newcommand{\OracleRelation}{\Relation^{\MathScriptStyle \Oracle}}
\newcommand{\ProverCount}{p}

\newcommand{\IOPRoundComplexity}{\mathsf{r}_{\scriptscriptstyle \IOP}}

\newcommand{\IOPOracleQueryComplexity}{\mathsf{q}_{\MathScriptStyle \Oracle}}

\newcommand{\VerifierPre}{\ver_{\MathScriptStyle \mathrm{pre}}}
\newcommand{\VerifierPost}{\ver_{\MathScriptStyle \mathrm{post}}}
\newcommand{\VerifierState}{\mathsf{st}_{\MathScriptStyle \ver}}
\newcommand{\IOPVerifierState}{\mathsf{st}_{\MathScriptStyle \IOPVerifier}}

\newcommand{\IOPVerifierPreTime}{t_{\MathScriptStyle \IOPVerifierPre}}
\newcommand{\SoundnessParameter}{\sigma}
\newcommand{\MIPSoundnessError}{\epsilon_{\scriptscriptstyle{\mathsf{MIP}}}}
\newcommand{\RIPSoundnessError}{\epsilon_{\scriptscriptstyle{\mathsf{RIP}}}}
\newcommand{\RIPRoundComplexity}{\mathsf{r}_{\scriptscriptstyle \mathsf{RIP}}}
\newcommand{\RIPQueryCount}{\mathsf{q}_{\MathScriptStyle \pro}}
\newcommand{\RIPCommunicationComplexity}{\mathsf{cc}}
\newcommand{\RIPOracleQueryComplexity}{\mathsf{q}_{\MathScriptStyle \Oracle}}
\newcommand{\RIPVerifierPreTime}{t_{\MathScriptStyle \VerifierPre}}
\newcommand{\RIPVerifierTime}{t_{\MathScriptStyle \ver}}
\newcommand{\RIPProverTime}{t_{\MathScriptStyle \pro}}
\newcommand{\IOP}{\mathsf{IOP}}
\newcommand{\IOPProver}{\mathcal{P}}
\newcommand{\IOPVerifier}{\mathcal{V}}
\newcommand{\IOPVerifierPre}{\IOPVerifier_{\MathScriptStyle \mathrm{pre}}}
\newcommand{\IOPVerifierPost}{\IOPVerifier_{\MathScriptStyle \mathrm{post}}}
\newcommand{\IOPProverTime}{t_{\MathScriptStyle \IOPProver}}
\newcommand{\IOPVerifierTime}{t_{\MathScriptStyle \IOPVerifier}}
\newcommand{\IOPSoundnessError}{\epsilon_{\scriptscriptstyle{\IOP}}}
\newcommand{\IOPProof}{\pi}
\newcommand{\IOPProofLength}{\ell}

\newcommand{\Alphabet}{\Sigma}
\newcommand{\Malicious}[1]{\widetilde{#1}}
\newcommand{\probability}[1]{\mathrm{Pr}\left[#1\right]}

\newcommand{\Field}{\mathbb{F}}
\newcommand{\MultilinearDimension}{h}
\newcommand{\EvaluationTable}{v}
\newcommand{\Transcript}{z}
\newcommand{\TranscriptLength}{N}

\newcommand{\EncodedWitnessLength}{\TranscriptLength'}

\newcommand{\BooleanPoint}{b}

\newcommand{\ClaimBit}{\zeta}
\newcommand{\TranscriptFunction}{\operatorname{Tr}}

\newcommand{\MLE}[1]{\widehat{#1}}

\newcommand{\EvaluationPoint}{r}

\newcommand{\CandidateTable}{u}

\newcommand{\PositionList}{\mathcal{I}}

\newcommand{\RestrictTo}[2]{#1|_{\MathScriptStyle #2}}
\newcommand{\DecodedBits}{\ClaimBit}

\newcommand{\SampleCount}{t}
\newcommand{\OraclePairSize}{w_{\scriptscriptstyle \Oracle}}

\newcommand{\RepetitionCount}{R}

\newcommand{\PrivateInput}{w}
\newcommand{\PrivateInputSize}{\InstanceSize_{\MathScriptStyle \PrivateInput}}

\newcommand{\ExternalVerifier}{\mathbf{D}}
\newcommand{\TestProver}{\IOPProver_{\MathScriptStyle \mathrm{dec}}}
\newcommand{\EWProver}{\IOPProver_{\MathScriptStyle \mathrm{ew}}}
\newcommand{\EWVerifier}{\IOPVerifier_{\MathScriptStyle \mathrm{ew}}}
\newcommand{\MaliciousEWProver}{\Malicious{\IOPProver}_{\MathScriptStyle \mathrm{ew}}}

\newcommand{\Code}{\mathcal{C}}
\newcommand{\Codeword}{c}
\newcommand{\RelaxedDecoder}{\mathsf{Dec}}
\newcommand{\DecoderQueryCount}{\mathsf{q}_{\MathScriptStyle \mathrm{dec}}}
\newcommand{\DecoderRun}[1]{\DecodedBits\leftarrow\RelaxedDecoder^{\MathScriptStyle #1}(\PositionList)}
\newcommand{\MaliciousTestProver}{\Malicious{\IOPProver}_{\MathScriptStyle \mathrm{dec}}}
\newcommand{\EWEncoding}{\mathsf{Enc}}
\newcommand{\EWMessageAlphabet}{\Gamma}

\newcommand{\EWValidCode}{\mathcal{C}_{\MathScriptStyle \Instance}}

\newcommand{\ExtensionError}{\epsilon_{\scriptscriptstyle{\mathrm{dec}}}}
\newcommand{\EWProximity}{\delta}
\newcommand{\EWDistanceBound}{\delta_{\EWEncoding}}
\newcommand{\EWDistance}{\Delta}
\newcommand{\EWLength}{k}
\newcommand{\EWDecision}{b}

\newcommand{\EWIOPRun}[1]{(\IOPProof_{\MathScriptStyle 1},\EWDecision)\leftarrow\langle #1,\IOPVerifier\rangle(\Instance)}

\newcommand{\EWTestRun}[2]{\DecodedBits\leftarrow\langle #1,\ExternalVerifier^{\MathScriptStyle #2}\rangle(\PositionList)}

\newcommand{\EqPoly}{\mathsf{eq}}

\newcommand{\IOPQueryCount}{\mathsf{q}_{\MathScriptStyle \IOPProof}}
\newcommand{\ProofQuery}{q}
\newcommand{\VerifierMessage}{\rho}
\newcommand{\FixedProofString}{s}

\title{Efficient Scalable Oversight via Rewinding}
\author{
Ziyi Guan\thanks{MIT. \texttt{ziyiguan@mit.edu}.} 
    \and 
Yael Tauman Kalai\thanks{MIT and RESI. \texttt{tauman@mit.edu}.} 
    \and 
Ron Rothblum\thanks{Succinct and Technion. \texttt{rothblum@gmail.com}.}
    \and
Zoe Xi\thanks{MIT. \texttt{zoexi@mit.edu}.}
}
\date{}

\begin{document}

\maketitle

\begin{abstract}
Scalable oversight asks how a weak verifier can check the outputs of computations performed by more powerful AI systems. One proposed approach uses relativizing interactive proofs, in which a weak verifier checks the output of a complex oracle-aided computation by interacting with one or more powerful provers. 
A prominent example is debate, which is highly efficient but assumes that one of two powerful AI models, or provers, is truthful. For certain classes of oracle-aided computations, this assumption can be removed, but at substantially greater computational cost.

In this work, we introduce \emph{rewindable interactive proofs} (RIPs), a new model of relativizing interactive proofs for scalable oversight. In an RIP, the verifier may rewind the prover to an earlier state of the interaction and continue from there. This is possible when the prover is an AI model, since the verifier can save a snapshot of the model's state and later restore it. We construct a relativizing RIP for computations whose outputs are robust to changing a small fraction of oracle answers. Our construction is more efficient than prior relativizing proofs that do not assume a truthful prover, and it uses no cryptography.

In our construction the honest prover runs in time linear in that of the original computation, and the verifier, after a one-time preprocessing step, runs in time nearly linear in the input length (which can be much smaller than the computation itself). We implement our protocol, report a preliminary MNIST experiment, and show that an RIP can be faster than a state-of-the-art deployed proof system.

RIPs may be of independent interest. Rewinding is possible whenever the verifier controls the prover's machine, for example when the prover runs on trusted hardware or in a virtual machine whose state can be checkpointed.
\end{abstract}

\newpage
\tableofcontents
\newpage

\section{Introduction}\label{sec:intro}
As AI capabilities grow rapidly, a critical problem is: how can we
ensure that AI systems remain safe and behave as intended? One major
concern is that, as we are training AI models to perform increasingly
complex tasks, we may no longer be able to \emph{efficiently} verify
the accuracy of their outputs.  As a result, we may not be able to
verify that they are behaving in a way that aligns with our
intentions. This problem of providing accurate training supervision
for increasingly powerful AI models is known as \emph{scalable
  oversight}.

Verifiable computation has been studied for decades in cryptography
and computational complexity. This research has led to proof systems
that allow a powerful but untrusted ``prover'' (in our setting, a
powerful AI model) to convince a weak ``verifier'' (a weaker AI model
or a human) that the output of a hard computation is correct
\cite{mip1988,feige1997making,arora1998probabilistic,arora1998proof,goldwasser2015delegating,rothblum2013interactive,ben2016interactive,reingold2016constant}.

In this work, we observe that scalable oversight has some unique
characteristics that make existing approaches either inapplicable or
less efficient than they could be. First, the prover is not an external party
but rather an AI model whose execution the verifier controls. In particular,
the verifier can \emph{rewind} the model to an earlier state and continue the interaction without the model remembering what happened in previous runs. To capitalize on this, we consider a new proof-system model which we call
\emph{rewindable interactive proofs} (RIPs),\footnote{Not to be
  confused with another RIP. Our provers are merely rewound, never
  harmed.} in which the verifier is allowed to rewind the prover in this way. This model is closely related to previous proof systems studied in the literature such as MIPs \cite{mip1988} and PCPs \cite{feige1997making,arora1998probabilistic,arora1998proof}, but we show that it can have significant efficiency gains over them.

Second, classical results in proof systems do not apply directly to our
setting, because the computations we wish to verify may involve
functions that we can't see the inner workings of---such a function can
be used to represent human judgment or query access to an external
database, for example. We model such ``black-box'' functions as \emph{oracles}, and
say that a proof system \emph{relativizes} if its guarantees hold with
respect to any oracle. Relativization is believed to be essential for
AI safety purposes, because we do not know what sorts of capabilities
AI may have in the future. The UK AISI team even writes that ``in order
to be relevant to AI safety, an interactive proof result must
relativize'' \cite{irving2025relativise}. The difficulty is that
classical results in proof systems, including the celebrated
$\mathsf{IP} = \mathsf{PSPACE}$ and $\mathsf{MIP} = \mathsf{NEXP}$
theorems, do not relativize \cite{chang1994random}, and moreover
results show that efficient verification is impossible for
arbitrary oracle-aided computations \cite{fortnow1988interactive,fortnow1994power,hartmanis1992relativization,fortnow1994role,chiesa2020impossibility,DBLP:conf/tcc/BarbaraCG25}.

For a concrete example of where scalable oversight is needed,
consider the scenario of training an LLM to determine which laws a
company has violated based on hundreds of pages of records. The LLM outputs one bit for
each law (representing yes or no), and now we need to verify whether
these bits are correct or not. The straightforward way of doing this
would be to have a human expert carefully read through the documents,
but this seems unreasonably expensive, especially to produce just one
training label. In a scenario like this, we would like to have a more efficient training protocol, especially one that is more efficient in its use of human judgments. As another example, we may consider a scenario where
we train an LLM to answer questions using a large public dataset, such
as the web; we would like to be able to verify that an output of this LLM is correct efficiently, making few queries to the dataset.

Various approaches to scalable oversight have been proposed in
the literature \cite{christiano2018supervising, leike2018scalable,
  irving2018ai}, including some based on relativizing proof
systems. One prominent example of such an approach is debate, which
was introduced by Irving, Christiano, and Amodei
\cite{irving2018ai}. In an AI debate, there are two powerful AI
models, or provers, who claim different outputs, and the provers
debate each other to try to convince a verifier that their output is
the correct one. The verifier is extremely efficient, or
\emph{succinct}, meaning that it runs in much less time and makes far
fewer oracle queries than it would need to run the computation on its
own. The authors show that $\mathsf{DEBATE}$, the class of problems
which admit a debate protocol, is powerful: in the language of
computational complexity theory, $\mathsf{DEBATE} = \mathsf{PSPACE}$
(where $\mathsf{PSPACE}$ is the class of problems solvable in
polynomial space). Moreover, this result relativizes.

However, debate makes the strong assumption that \emph{one of the
  provers is truthful}, i.e., one of the AI models is guaranteed to
argue for the correct answer. Intuitively, this means that the
verifier has one of the powerful AI models on his side. But scalable
oversight is especially important precisely when we do not know
whether we can trust any AI model. Recently, Chen, Kalai, and Xi \cite{chen2026avoid} sought to
remove this assumption by constructing ``doubly-efficient''
\emph{interactive proofs} \cite{goldwasser2015delegating}. In an
interactive proof \cite{goldwasser2019knowledge, babai1985trading}, a
prover \emph{interacts} with a verifier to convince him of a claim; a
doubly-efficient one has an efficient prover as well as an efficient
verifier. Since this is impossible for arbitrary oracle computations, \cite{chen2026avoid} instead give protocols for a natural restricted setting
where the computation is \emph{robust}, meaning that the output does not
change even if a small fraction of answers to oracle queries are
modified. However, these protocols still require substantially more
work from the verifier than debate. A natural question, then, is:

\begin{center}
    \emph{Can we obtain a more efficient relativizing interactive proof without assuming any prover is truthful?}
\end{center}

Motivated by this question, we give a positive answer for robust oracle computations in the RIP model: by exploiting the verifier's ability to \emph{rewind} the prover, we obtain strong efficiency guarantees without assuming that any prover is truthful.

\subsection{Our results}

We introduce and formally define rewindable interactive proofs, and
construct a relativizing RIP for robust oracle-aided computations,
with a succinct verifier and a linear-time prover. We also implement
our protocol and evaluate its performance experimentally.

\paragraph{The model.}
In an interactive proof, a prover and a verifier exchange messages,
after which the verifier accepts or rejects. The proof must be
\emph{complete}, meaning that the honest prover convinces the verifier
of a correct output, and \emph{sound}, meaning that no cheating prover
can convince it of an incorrect output, except with small probability.
An RIP is an interactive proof in which the verifier has one extra
power: at any point, it may rewind the prover to an earlier state of
the interaction and continue from there, as one would restore an AI
model from a save point (\cref{sec:rip-definition} gives the formal
definition). Completeness is as usual: the honest prover must convince
the verifier, now across whatever rewinds the verifier performs.
Soundness, on the other hand, asks less than it does of a standard proof system. A
standard proof system must be sound against every cheating prover,
however it behaves; an RIP need only be sound against provers that the
verifier can rewind, whose answer after a rewind cannot depend on
anything the verifier did after the restored state. This is a weaker
guarantee, and it is what makes the model powerful: the verifier can
question the prover about one state as many times as it likes, and
the prover never learns the other questions. We emphasize that soundness is violated if a cheating prover has access to external storage that is not rewound.

\paragraph{A relativizing RIP for robust computations.}
Following Chen, Kalai,
and Xi \cite{chen2026avoid}, we model the computation as an oracle Boolean circuit and call it $\eps$-robust if its
output does not change when up to an $\eps$-fraction of the oracle's
answers are modified. Our main result is a relativizing RIP for
verifying the output of any robust oracle circuit, with a succinct
verifier and a prover that runs in linear time.

\begin{theorem}[Informal; see \cref{thm:robust-rip-parameters}]
\label{thm:robust-mip-informal}
  Let $\OracleCircuit$ be an oracle circuit with $\InstanceSize$ input
  bits, $\OutputSize$ output bits, and $\CircuitSize$ gates, whose
  oracle queries have polylogarithmic length. After a one-time
  preprocessing step taking
  $\SoftO{\InstanceSize+\OutputSize+\CircuitSize}$ time, there is a
  relativizing RIP for verifying the output of $\OracleCircuit$ on any
  input with respect to which it is $\eps$-robust, with the following
  guarantees:
\begin{itemize}[noitemsep]
\item The honest prover always convinces the verifier of a correct
  output, while a cheating prover (even a computationally unbounded
  one) can convince the verifier of an incorrect output with
  probability at most $1/3$.
\item The verifier runs in $\SoftO{\InstanceSize+\OutputSize+1/\eps}$
  time and makes $\BigO{1/\eps}$ oracle queries.
\item The honest prover runs in
  $\BigO{\InstanceSize+\OutputSize+\CircuitSize}+\SoftO{1/\eps}$ time.
\item The number of rounds is
  $\BigO{\log(\InstanceSize+\OutputSize+\CircuitSize)}$ and the total
  communication is $\SoftO{1/\eps}$.
\end{itemize}
  Here $\SoftO{\cdot}$ hides factors polylogarithmic in
  $\InstanceSize+\OutputSize+\CircuitSize$, and running times count
  field operations.
\end{theorem}

In the \emph{preprocessing} step, the verifier reads the description
of the circuit and stores a representation of it, which it reuses in
every evaluation of the circuit; the cost is thus amortized over all
evaluations, which suits our setting, where a fixed circuit is
evaluated on many inputs. Some such step is necessary to achieve
online verification time sublinear in the size of a general circuit,
and preprocessing is standard in efficient proof systems
\cite{chiesa2020marlin,setty2020spartan,chen2022hyperplonk}; see
\cref{rem:iop-preprocessing}.

\paragraph{Comparison with \cite{chen2026avoid} and debate.}
Compared with the protocols of \cite{chen2026avoid}, after the preprocessing step our verifier's running time depends
only on the lengths of the input and output and on the robustness
parameter, instead of on the size or depth of the circuit, and its
number of oracle queries depends only on the robustness parameter;
compared with their succinct protocol, ours is statistically sound,
uses no cryptography, and has a linear-time rather than a
polynomial-time prover. Unlike debate, our protocol does not assume
that a prover is truthful.

\paragraph{Why not concretely efficient SNARGs?}
The proof systems deployed in practice, notably in blockchains
\cite{bensasson2014zerocash,DBLP:journals/ftsec/Thaler22}, are SNARGs,
succinct non-interactive arguments. Why not take one off the shelf?
First, off-the-shelf SNARGs do not handle oracle-aided computations,
even robust ones. Second, their soundness rests on unproven
cryptographic assumptions, from standard ones such as
collision-resistant hash functions to more aggressive ones chosen for
concrete efficiency. As AI brings such assumptions under increasing
scrutiny (recent AI-discovered results, still under review, claim to
refute the Exponential Time Hypothesis
\cite{openai2026eth,anthropic2026eth} and, assuming them, give a
subexponential algorithm for the elliptic-curve discrete logarithm
problem \cite{akshima2026ecdl}), it is unclear whether this is a
prudent tradeoff for scalable oversight. RIPs address both concerns:
they handle robust oracle-aided computations without cryptographic
assumptions and are sound even against computationally unbounded
provers. Perhaps surprisingly, dispensing with cryptography can also
\emph{improve} prover efficiency: rewinding avoids the cost of
cryptographic commitments (\cref{rem:complexity-preserving}), and our
second experiment demonstrates the resulting speedup.

\paragraph{Experiments.}
We complement our theoretical result by implementing the protocol and
running two experiments. The first is a proof-of-concept experiment
inspired by the MNIST debate experiment of Irving, Christiano, and
Amodei~\cite{irving2018ai}: we run the protocol once, at soundness
error $2^{-40}$, on robust oracle circuits with between $2^{8}$ and
$2^{26}$ oracle gates, whose oracle is the debate experiment's judge,
a sparse MNIST classifier. Across this range the verifier makes a
median of about $76$ oracle queries, against up to $2^{26}$ needed to
evaluate the largest circuit directly, and its online running time
falls below that of direct evaluation from about seven million oracle
gates on, to an eighth of it at the largest size.

For our second experiment, we consider Flock \cite{bunz2026flock}, a
state-of-the-art SNARG built from an interactive oracle proof (IOP)
\cite{ben2016interactive,reingold2016constant} (see
\cref{sec:technical-overview}) for batch evaluation of the BLAKE3 hash
function. To study the efficiency gains of rewinding, we
compile Flock's IOP into an RIP and find that its prover runs $1.8$ to
$2.1\times$ faster than Flock's own SNARG prover
(\cref{sec:impl-blake3}).

\subsection{Related work}

\paragraph{Relativizing proof systems.} Relativization has been
studied extensively as a barrier in computational complexity theory,
and notably the classical results on interactive proofs, e.g., the
celebrated $\mathsf{IP} = \mathsf{PSPACE}$ and
$\mathsf{MIP} = \mathsf{NEXP}$ theorems, do not relativize
\cite{chang1994random}. Additionally on the negative side, results show that
efficient verification is impossible for arbitrary oracle-aided
computations (even when the security guarantee is relaxed to hold only
against efficient dishonest provers) \cite{fortnow1988interactive,fortnow1994power,hartmanis1992relativization,fortnow1994role,chiesa2020impossibility,DBLP:conf/tcc/BarbaraCG25}. Our work circumvents these barriers by
restricting to the setting of robust oracle computation. On the
positive side, prior work has shown that
$\mathsf{IP} = \mathsf{PSPACE}$ and $\mathsf{MIP} = \mathsf{NEXP}$ do
algebrize, meaning that they relativize if the proof system is also
given a low-degree extension of the oracle
\cite{aaronson2009algebrization}, and there has also been work on
constructing relativizing succinct cryptographic proof systems in some
structured oracle models
\cite{chiesa2010proof,chen2022relativized,chen2023arithmetized,chiesa2024straightline,chen2026avoid}.

\paragraph{Interactive proofs for scalable oversight.} There has been a recent
line of work studying interactive proofs as an approach to scalable
oversight, starting with a paper by Irving, Christiano, and Amodei on
``AI safety via debate'' \cite{irving2018ai}. Their model assumes that the debating
provers are computationally unbounded; later, Brown-Cohen, Irving, and
Piliouras introduced doubly-efficient debate, where the provers are
guaranteed to be efficient \cite{brown2024scalable}. More recent works on debate have focused
on its limitations, e.g., the ``obfuscated arguments problem,'' where a
lying debating prover can make it very hard for the other prover to find the flaw in her
argument \cite{barnes2020obfuscated, brown2025avoiding, buhl2025alignment}. Very recently, Chen, Kalai, and Xi initiated the study of
relativizing interactive proofs for scalable oversight without a
truthful prover, constructing protocols with a succinct verifier and a polynomial-time
prover for certain classes of oracle computations \cite{chen2026avoid}. Our work considers the setting of robust computation
that they introduce; we leverage rewinding to improve the
efficiency of their protocols.

\paragraph{Interactive proofs for trustworthy machine learning.} There
is also a line of work on applying interactive proofs to problems of
trust in machine learning \cite{amit2024models, anil2021learning,
  hammond2024neural, kirchner2024prover,
  wäldchen2024interpretabilityguaranteesmerlinarthurclassifiers}. In
particular, Amit, Goldwasser, Paradise, and Rothblum
\cite{amit2024models} introduce self-proving models, which are trained
to provide both an answer and an interactive proof that the answer is
correct; their focus is on training models to produce proofs for
computations that already admit interactive proofs, whereas we
construct new interactive proofs for oracle-aided computation.

\UseScriptScriptStyle
\UseDefinitionStyleForTheorems
\section{Technical Overview}\label{sec:technical-overview}
Here, let us briefly outline the key ideas behind our construction.

We model the computation that the verifier wishes to verify as an
oracle Boolean circuit $C^O$, i.e., a Boolean circuit that may contain
gates that make calls to an arbitrary oracle
$O:\{0,1\}^*\to\{0,1\}$. We assume that the oracle circuit is
$\eps$-robust with respect to the verifier's input, meaning that its
output on this input does not change if at most an $\eps$-fraction of
answers to its oracle queries are changed (see \cref{defn:robustness}
for a formal definition). Given an oracle circuit $C^O$, an input $x$
for which $C^O$ is $\eps$-robust, a claimed output $a$, and access to
$O$, the prover and the verifier engage in a rewindable interactive
proof (RIP, \cref{sec:rewindable}) that $C^O(x) = a$. Recall that the verifier may rewind the
prover to any earlier state of the interaction and continue from there.

Our goal is to design a protocol that is efficient enough to use in
practice. Most practical proof systems are built from an interactive
oracle proof (IOP) \cite{ben2016interactive,reingold2016constant},
which is an interactive proof where the prover sends long
messages but the verifier reads only a few bits of each message,
because IOPs admit very efficient provers and verifiers. We first
construct a relativizing IOP and then turn it into a relativizing
RIP that inherits the IOP's efficiency.

\paragraph{A relativizing IOP (\cref{sec:relativized-iops}).} To start, a natural strategy is to try
to turn the oracle circuit $C^O$ into a circuit without oracle gates
and then use standard IOPs. To ``de-oraclize'' the circuit, suppose
that the prover provides a string $w$ containing a claimed answer for
each oracle gate. We can then define a circuit $C_w$ without oracle
gates, which is $C^O$ except with each oracle gate replaced by the
corresponding answer in $w$. The basic idea is to have the prover send
the string $w$ in the first round of the IOP and then prove that
$C_w(x) = a$. Then the verifier can check that the answers in $w$ are
correct with respect to $O$.

The verifier cannot check that all the answers in $w$ are correct,
since this would require making all the oracle calls needed in order
to run the computation, which he cannot afford. But the verifier does
not actually need to do this, since we assume $C^O$ is $\eps$-robust:
if $C^O(x)\neq a$, then any $w$ satisfying $C_w(x) = a$ must contain
more than an $\eps$-fraction of incorrect oracle answers, so the
verifier can catch an incorrect oracle answer with good probability by
sampling $O(1/\eps)$ answers in $w$.

To obtain stronger efficiency guarantees, instead of $w$ itself, we
have the prover send an encoding of a witness consisting of $w$
together with a ``transcript'' of evaluating $C_w$ on $x$. The
prover and verifier then run an IOP to check that this encoding
is close to the encoding of a valid witness for $C_w(x) = a$.
The advantage of using an encoding is that the verifier can later
recover a few bits of this same witness, with the prover's help,
without reading the entire witness. In particular, the verifier
can sample a few oracle gates, recover the claimed answer in $w$
for each oracle gate and the corresponding query using the
transcript, and check that the pair is correct using $O$.
To achieve this, we construct a variant of an IOP that we call
an encoded-witness IOP (\cref{sec:ewiop}).

\paragraph{Relativizing IOP $\rightarrow$ relativizing RIP (\cref{sec:iop-to-rip}).} Now we
turn our relativizing IOP into a relativizing RIP. The idea is similar to the one behind the standard proof for
turning a (non-relativizing) IOP into an MIP
\cite{ben2016interactive} (see \cref{app:mip} for MIPs and their relation to RIPs). We have the RIP verifier simulate the IOP
verifier and the RIP prover simulate the IOP prover, except that the
``IOP'' prover computes her long messages but does not send them. If
the verifier simply asked the prover for the bits he wants to read,
she could answer adaptively, e.g., her answer to the verifier's second
query to some message could depend on his first query. Instead,
whenever the verifier wants to read a bit in some message, he rewinds
the prover to her state right after she computed that message and
asks her for the bit. Her answer then cannot depend on his other
queries, so she cannot do better than an IOP prover who fixes each
message before the verifier queries it, and the soundness error is
that of the IOP. Rewinding thus allows the RIP to be as efficient as
the IOP.

\paragraph{Experiments (\cref{sec:implementation}).} We implement our protocol and run two
experiments. The first evaluates it on robust oracle circuits and
measures what the verifier saves over evaluating the circuit
directly. The second compares it with an existing proof system. Since
no relativizing protocol has an implementation, we compare with a
state-of-the-art proof system for an oracle-free computation, which is
built from an IOP: compiling the same IOP into an RIP instead measures
what rewinding saves.

The first experiment is a proof of concept inspired by the MNIST
debate experiment of Irving, Christiano, and
Amodei~\cite{irving2018ai}. We use a sparse MNIST classifier,
which predicts a digit from six revealed nonzero pixels of an MNIST
image, to define our oracle. Given a ``sparse view'' of an MNIST image
and a candidate digit, the oracle outputs whether the classifier
predicts that digit. Our circuits query this oracle on many random
sparse views of an image and output whether a majority of the views
vote for the candidate digit.

We run the protocol once, at soundness error $2^{-40}$, on 50 MNIST
test images while increasing the number $q_C$ of oracle gates from
$2^{8}$ to $2^{26}$. The robustness parameter remains nearly constant
across these circuit sizes, and the median number of oracle queries
made by the verifier stays between $74$ and $80$. At the largest
circuit size, the verifier makes $840{,}000\times$ fewer oracle
queries than are needed to evaluate the circuit directly. The
verifier's online running time is also sublinear in the circuit size,
growing substantially more slowly than direct evaluation: it falls
below direct evaluation at about seven million oracle gates, and at
$q_C=2^{26}$ is $0.12\times$ the time required to evaluate the circuit
directly. The prover, which also generates the witness, takes about
$60\times$ the time of direct evaluation at the largest sizes.

For the second experiment we use Flock \cite{bunz2026flock}, a proof
system for batch evaluation of an ordinary Boolean circuit with no
oracle gates, namely the BLAKE3 hash function. Flock is built from
an IOP, which is not by itself a realistic proof system, since the
prover sends long strings; Flock compiles it into a SNARG using
cryptography. We compile the same IOP into an RIP instead, which saves
the cryptographic compilation: the prover no longer commits to its
proof strings, and since a cheating prover gets one attempt per
execution rather than unlimited offline retries, the IOP can run at a
lower security level and over a smaller field. On instances of size $2^{22}$ to $2^{34}$, our prover
runs $1.8$ to $2.1\times$ faster than Flock's prover
(\cref{sec:impl-blake3}).

\section{Preliminaries}\label{sec:prelims}
Here we introduce some definitions and tools that we use.

\begin{definition}[Language]\label{defn:language}
  A \defn{language} $\Language\subseteq\bits^{\MathScriptStyle *}$ is a set of binary strings. We think of a language as a decision problem: given an \defn{instance} $\Instance\in\bits^{\MathScriptStyle *}$, we want to determine whether $\Instance\in\Language$.

\end{definition}

\begin{definition}[Relation]\label{defn:relation}

  A \defn{relation} 
$\Relation\subseteq\bits^{\MathScriptStyle *}\times\bits^{\MathScriptStyle *}$ is a set of pairs $(\Instance,\Witness)$, where $\Instance$ is an instance and $\Witness$ is a
  \defn{witness} for $\Instance$. The language associated with $\Relation$ is
  $
    \mathcal{L}\coloneq\{\Instance: \exists \Witness (\Instance,\Witness)\in \Relation\}.
  $

\end{definition}

\begin{definition}[Oracle and oracle access]\label{defn:oracle}
  An \defn{oracle} is a fixed function $\Oracle:\{0, 1\}^*\to\{0, 1\}$;
  in particular, its answer to a query does not depend on earlier
  queries. We
  write that an algorithm has \defn{oracle access} to $\Oracle$ if it
  can send a query $\OracleQuery\in\{0, 1\}^*$ and receive the answer
  string $\Oracle(\OracleQuery)$ in one timestep.
\end{definition}

\begin{definition}[Relativized language]\label{defn:relativized-language}

  For an oracle $\Oracle$, a \defn{relativized language} $\OracleLanguage$ is a language whose membership may depend on $\Oracle$.

\end{definition}

\begin{definition}[Relativized relation]\label{defn:relativized-relation}

  For an oracle $\Oracle$, a \defn{relativized relation} $\OracleRelation$ is a relation whose membership may depend on $\Oracle$. Its associated relativized language is $\mathcal{L}_{R^O}$.

\end{definition}

\begin{definition}[Equality polynomial]\label{defn:equality-polynomial}

  Let $\Field$ be a finite field and $\MultilinearDimension\ge0$ an integer. The \defn{equality polynomial} $\EqPoly:\Field^{\MathScriptStyle \MultilinearDimension}\times\Field^{\MathScriptStyle \MultilinearDimension}\to\Field$ is
  \[
    \EqPoly(\EvaluationPoint,e)\coloneq\prod_{\MathScriptStyle i=1}^{\MathScriptStyle \MultilinearDimension}\bigl(\EvaluationPoint_{\MathScriptStyle i}e_{\MathScriptStyle i}+(1-\EvaluationPoint_{\MathScriptStyle i})(1-e_{\MathScriptStyle i})\bigr).
  \]
  For $\BooleanPoint,e\in\bits^{\MathScriptStyle \MultilinearDimension}$, $\EqPoly(\BooleanPoint,e)$ is $1$ if $\BooleanPoint=e$ and $0$ otherwise.

\end{definition}

\begin{definition}[Multilinear extension]\label{defn:multilinear-extension}

  Let $\Field$ be a finite field and $\MultilinearDimension\ge 0$ an integer. We interpret every vector $\EvaluationTable\in\Field^{\MathScriptStyle 2^{\MathScriptStyle \MultilinearDimension}}$ as a function $\EvaluationTable:\bits^{\MathScriptStyle \MultilinearDimension}\to\Field$. The \defn{multilinear extension} of $\EvaluationTable$ is the unique polynomial $\MLE{\EvaluationTable}:\Field^{\MathScriptStyle \MultilinearDimension}\to\Field$ of degree at most one in each variable satisfying $\MLE{\EvaluationTable}(\BooleanPoint)=\EvaluationTable(\BooleanPoint)$ for every $\BooleanPoint\in\bits^{\MathScriptStyle \MultilinearDimension}$. Explicitly,
  \[
    \MLE{\EvaluationTable}(\EvaluationPoint)
    =\sum_{\MathScriptStyle \BooleanPoint\in\bits^{\MathScriptStyle \MultilinearDimension}}
      \EvaluationTable(\BooleanPoint)\,\EqPoly(\EvaluationPoint,\BooleanPoint),
  \]
  where $\EqPoly$ is the equality polynomial (\cref{defn:equality-polynomial}).
  For a binary string of length $2^{\MathScriptStyle \MultilinearDimension}$, we view its bits as elements of $\Field$ and use the same definition. For a vector or string of any other length, the multilinear extension is that of its zero-padding to the next power-of-two length $2^{\MathScriptStyle \MultilinearDimension}$.

\end{definition}

\subsection{Oracle-aided computation and robustness}

Following \cite{chen2026avoid}, we model the computation that the verifier verifies as an oracle Boolean circuit. Informally, this is a diagram that shows how to obtain an output string from an input string by applying some sequence of OR ($\lor$), AND ($\land$), and NOT ($\neg$) operations together with calls to an oracle. More formally, we have:

\begin{definition}[Boolean circuit]\label{defn:boolean-circuit}

  An $\InstanceSize$-bit input, $\OutputSize$-bit output Boolean circuit is a directed acyclic graph with $\InstanceSize$ input vertices and an ordered list of $\OutputSize$ output vertices. Input vertices have no incoming edges; every other vertex is a \defn{gate} labeled $\lor$ or $\land$ with two incoming edges, or $\neg$ with one. Any vertex may be an output. A circuit $\Circuit$ is given by its \defn{description} $\langle\Circuit\rangle$, a binary string that lists its gates in a fixed order, each with its label and its incoming edges, followed by the list of output vertices; we refer to gates by their position in this list.

  On input $\CircuitInput\in\bits^{\MathScriptStyle \InstanceSize}$, the value of the $i$-th input vertex is the $i$-th bit of $x$, and each gate applies its operation to the values of its \defn{children}, i.e., the vertices with edges going into the gate. The output of the circuit on $x$ consists of the values of the output vertices read in order. The \defn{size} of a circuit is its number of gates.

\end{definition}

\begin{definition}[Oracle Boolean circuit]\label{defn:oracle-circuit}

  For an oracle $\Oracle:\bits^{\MathScriptStyle *}\to\bits$, an
  oracle Boolean circuit $\OracleCircuit$ is a Boolean circuit
  that may also contain \defn{oracle gates}. Each oracle gate has an
  arbitrary number of incoming edges. We view the input to 
  an oracle gate as the string given by the values of its children
  read in order. The value of an oracle gate is $\Oracle$ applied to
  its input string. 

\end{definition}

\begin{definition}[Family of circuits]\label{defn:circuit-family}

  A family of circuits $\{\Circuit_{\MathScriptStyle \InstanceSize}\}_{\MathScriptStyle \InstanceSize\in\bbN}$ consists of one circuit $\Circuit_{\MathScriptStyle \InstanceSize}$ with an $\InstanceSize$-bit input and an $\OutputSize(\InstanceSize)$-bit output for each $\InstanceSize$. It \defn{computes} a function $\ComputedFunction:\bits^{\MathScriptStyle *}\to\bits^{\MathScriptStyle *}$ if $\Circuit_{\MathScriptStyle \InstanceSize}(\CircuitInput)=\ComputedFunction(\CircuitInput)$ for every $\InstanceSize$ and every $\CircuitInput\in\bits^{\MathScriptStyle \InstanceSize}$.

\end{definition}

\begin{definition}\label{defn:computation-transcript}

  Let $\OracleCircuit$ be an oracle circuit of size $\CircuitSize$. We define the computation transcript on input $\CircuitInput$ to be the string that lists the values of all the gates in $C$ when run on input $x$ in order (using $\Oracle$ to evaluate oracle gates), and denote the computation transcript by
  \[
    \Transcript=\TranscriptFunction_{\MathScriptStyle \OracleCircuit}(\CircuitInput)\in\bits^{\MathScriptStyle \CircuitSize}.
  \]
\end{definition}

\begin{definition}\label{defn:query-answer}
Suppose $\OracleCircuit$ has $\OracleGateCount$ oracle gates, and consider running $C$ on an input $x$. We use 
  \[
    \QueryAnswerTrace_{\MathScriptStyle \OracleCircuit}(\CircuitInput) = \QueryAnswerTrace=((\OracleQuery_{\MathScriptStyle 1},\OracleAnswer_{\MathScriptStyle 1}),\ldots,(\OracleQuery_{\MathScriptStyle \OracleGateCount},\OracleAnswer_{\MathScriptStyle \OracleGateCount}))
  \]
  for the string of queries and answers that $\OracleCircuit$ makes to $O$ on input $x$, where $\OracleQuery_i$ (resp. $\OracleAnswer_i$) is the input to (resp. output of) oracle gate $i$. Call $\QueryAnswerTrace$ a query-answer trace. We write $\QueryInputs{\QueryAnswerTrace}$ (resp. $\QueryAnswers{\QueryAnswerTrace}$) for the string of queries (resp. answers) in $\QueryAnswerTrace$. 

\end{definition}  

\begin{definition}[De-oraclized circuit]\label{defn:deoraclized}

  Let $\OracleCircuit$ be an oracle circuit with an input of length $\InstanceSize$ and $\OracleGateCount$ oracle gates. We say that its \defn{de-oraclized circuit} is the (non-oracle) circuit $\Circuit$ with input $(\CircuitInput,\PrivateInput)\in\bits^{\MathScriptStyle \InstanceSize}\times\bits^{\MathScriptStyle \OracleGateCount}$ obtained by replacing, for every $j\in[\OracleGateCount]$, oracle gate $j$ with an input vertex with value $\PrivateInput_{\MathScriptStyle j}$. 
\end{definition}

We next define robustness. Informally, the output of a robust oracle circuit should not change if at most a small fraction of answers to its oracle queries are incorrect. More formally, we define robustness as follows, adapting the definition that Chen, Kalai, and Xi use \cite{chen2026avoid}:

\begin{definition}[Robustness]\label{defn:robustness}

  Let $\OracleCircuit$ be an oracle circuit with an input of length $\InstanceSize$ and $\OracleGateCount$ oracle gates, and let $\Circuit$ be its de-oraclized circuit (\cref{defn:deoraclized}). For an input $(x, w)$, where $\CircuitInput\in\bits^{\MathScriptStyle \InstanceSize}$ and $\PrivateInput\in\bits^{\MathScriptStyle \OracleGateCount}$, and an index $j\in[\OracleGateCount]$, let $\OracleQuery_{\MathScriptStyle j}(\PrivateInput)$ denote the input to oracle gate $j$ in $\OracleCircuit$ when we run $\OracleCircuit$ on $\CircuitInput$ except with the value of each oracle gate replaced with the corresponding bit in $w$. For $\eps\in (0, 1]$, we say that $\OracleCircuit$ is \defn{$\eps$-robust} with respect to $\CircuitInput\in\bits^{\MathScriptStyle \InstanceSize}$ if, for every $\PrivateInput\in\bits^{\MathScriptStyle \OracleGateCount}$ satisfying
  \[
    \bigl|\{j\in[\OracleGateCount]:\PrivateInput_{\MathScriptStyle j}\ne\Oracle(\OracleQuery_{\MathScriptStyle j}(\PrivateInput))\}\bigr|\le\eps\OracleGateCount,
  \]
  we have that $\Circuit(\CircuitInput,\PrivateInput)=\OracleCircuit(\CircuitInput)$. 

\end{definition}

\begin{definition}[Circuit-satisfiability relation]\label{defn:csat-relation}
  The \defn{circuit-satisfiability relation} $\CSATRelation$ is the relation defined as follows. The instance is a triple $\Instance=(\langle\Circuit\rangle,\CircuitInput,\OutputValue)$, where $\langle\Circuit\rangle$ is the description of a Boolean circuit with $\InstanceSize+\PrivateInputSize$ input vertices, of which the first $\InstanceSize$ form the \defn{public input} and the remaining $\PrivateInputSize$ the \defn{private input}, an output of length $m$, and $\CircuitSize$ gates, $\CircuitInput\in\bits^{\MathScriptStyle \InstanceSize}$ is a public input, and $\OutputValue\in\bits^{\MathScriptStyle \OutputSize}$ is a claimed output. The witness is a pair $\Witness=(\PrivateInput,\Transcript)\in\bits^{\MathScriptStyle \PrivateInputSize}\times\bits^{\MathScriptStyle \CircuitSize}$ of a private input and a transcript. Then
  \begin{equation*}
    (\Instance,\Witness)\in\CSATRelation
    \quad\Longleftrightarrow\quad
    \Transcript=\TranscriptFunction_{\MathScriptStyle \Circuit}(\CircuitInput,\PrivateInput)
    \;\land\;\Circuit(\CircuitInput,\PrivateInput)=\OutputValue,
  \end{equation*}
  where $\TranscriptFunction_{\MathScriptStyle \Circuit}$ is the computation transcript of \cref{defn:computation-transcript}.
\end{definition}

\begin{definition}[Oracle-circuit-evaluation relation]\label{defn:cval-relation}
  Fix an oracle $\Oracle$. The \defn{oracle-circuit-evaluation relation} $\CvalRelation$ is the relativized relation (\cref{defn:relativized-relation}) defined as follows. The instance is a triple $\Instance=(\langle\OracleCircuit\rangle,\CircuitInput,\OutputValue)$, where $\langle\OracleCircuit\rangle$ is the description of an oracle circuit with an input of length $\InstanceSize$, an output of length $\OutputSize$, and $\CircuitSize$ gates, $\CircuitInput\in\bits^{\MathScriptStyle \InstanceSize}$ is an input, and $\OutputValue\in\bits^{\MathScriptStyle \OutputSize}$ is a claimed output. The witness is a transcript $\Transcript\in\bits^{\MathScriptStyle \CircuitSize}$. Then
  \begin{equation*}
    (\Instance,\Transcript)\in\CvalRelation
    \quad\Longleftrightarrow\quad
    \Transcript=\TranscriptFunction_{\MathScriptStyle \OracleCircuit}(\CircuitInput)
    \;\land\;\OracleCircuit(\CircuitInput)=\OutputValue,
  \end{equation*}
  where $\TranscriptFunction_{\MathScriptStyle \OracleCircuit}$ is the computation transcript of \cref{defn:computation-transcript}.
\end{definition}

\begin{remark}[Transcript as witness]\label{rem:cval-witness}
  The witness of \cref{defn:csat-relation} is a private input $\PrivateInput$ together with the entire transcript $\Transcript$, and the witness of \cref{defn:cval-relation} is the entire transcript. The transcript is not needed to define either relation, as it is determined by the instance and $\PrivateInput$ in \cref{defn:csat-relation} and by the instance alone in \cref{defn:cval-relation}. Including it nevertheless allows succinct verification, since the verifier can check an encoding of the witness at a few positions instead of evaluating the circuit (\cref{defn:ewiop,lem:ewiop-circuits,ewiop-to-reliop}).
\end{remark}

\subsection{Interactive oracle proofs}
\label{section:iop}

An interactive oracle proof (IOP), introduced in \cite{ben2016interactive, reingold2016constant}, lets a powerful prover convince a weak verifier that an instance has a valid witness by sending long proof strings that the verifier checks at a few locations. The parties may interact over several rounds.

\begin{definition}[Interactive oracle proof]
\label{defn:iop}

  An interactive oracle proof for a relation $\Relation$ is an interactive protocol between a probabilistic polynomial-time verifier $\IOPVerifier$ and a possibly computationally unbounded prover $\IOPProver$. The parties receive a common input $\Instance$, and the honest prover additionally receives a witness $\Witness$ with $(\Instance,\Witness)\in \Relation$.

  In round $i$, the prover sends a proof string $\IOPProof_{\MathScriptStyle i}$ over a finite alphabet $\Alphabet$, which is a finite field in our constructions. The verifier has oracle access (\cref{defn:oracle}) to every proof string sent so far.

  The protocol satisfies the completeness and soundness properties defined below.

\end{definition}

We write $\langle\IOPProver(\Witness),\IOPVerifier\rangle(\Instance)=1$ when the verifier accepts after interacting with the prover on common input $\Instance$. Note that we denote the IOP prover and verifier by $\IOPProver$ and $\IOPVerifier$, rather than by $\pro$ and $\ver$ as for rewindable interactive proofs (\cref{sec:rewindable}) and MIPs (\cref{app:mip}), to distinguish between the settings. The probabilities below are over the verifier's private random bits (we assume without loss of generality that the prover is deterministic). 

\begin{definition}[Completeness]
\label{definition:iop-completeness}

  An IOP for $\Relation$ has
  \defn{perfect completeness} if, for every $(\Instance,\Witness)\in \Relation$,
  the honest prover satisfies
  \[
    \probability{
      \langle\IOPProver(\Witness),\IOPVerifier\rangle(\Instance)=1
    }=1.
  \]

\end{definition}

\begin{definition}[Soundness]
\label{definition:iop-soundness}

  An IOP for $\Relation$ has
  \defn{soundness error} $\IOPSoundnessError$ if, for every
  $\Instance\notin \Language$ and every possibly computationally unbounded cheating prover $\Malicious{\IOPProver}$,
  \[
    \probability{
      \langle\Malicious{\IOPProver},\IOPVerifier\rangle(\Instance)=1
    }\le\IOPSoundnessError(|\Instance|).
  \]

\end{definition}

\begin{definition}[Relativizing IOP]\label{defn:relativized-iop}

  Fix an oracle $\Oracle$ and a relativized relation $\OracleRelation$. A \defn{relativizing IOP} for $\OracleRelation$ is an IOP for $\OracleRelation$ in which the verifier and prover, including a cheating prover, have black-box access to the same oracle $\Oracle$. The protocol \defn{relativizes} if its completeness, soundness, and efficiency guarantees hold for $\OracleRelation$ for every oracle $\Oracle$.

\end{definition}

\begin{remark}[Preprocessing]\label{rem:iop-preprocessing}
  An instance may have the form $\Instance=(\InstancePre,\InstancePost)$, where $\InstancePre$ is long and is the same for many instances. In our setting $\InstancePre$ is the description of a circuit and $\InstancePost$ is an input together with a claimed output (\cref{defn:cval-relation}). An IOP \defn{with preprocessing} has a verifier that is a pair $\IOPVerifier=(\IOPVerifierPre,\IOPVerifierPost)$. First $\IOPVerifierPre$ receives $\InstancePre$ and outputs a state $\IOPVerifierState$. Then $\IOPVerifierPost$ receives $\IOPVerifierState$ and $\InstancePost$ and interacts with the prover. The verifier reads $\InstancePre$ only in the preprocessing, which is done once and whose state is reused for every $\InstancePost$. Completeness and soundness are as in \cref{definition:iop-completeness,definition:iop-soundness}.

  For the verifier's online time to be sublinear in the size of a general circuit, we need to preprocess the circuit. Intuitively, without preprocessing the verifier must read the entire circuit, since two circuits that differ in one gate can produce different outputs on the same input. Preprocessing is standard in deployed proof systems \cite{chiesa2020marlin,setty2020spartan,chen2022hyperplonk}. Since we typically fix a circuit and evaluate it on many inputs, the one-time cost of preprocessing is amortized over all the evaluations. Preprocessing is unnecessary, however, for a circuit with a short description, since the verifier can then work from the description instead of reading the whole circuit \cite{goldwasser2015delegating,cormode2012practical,thaler2013time,xie2019libra}. It is also unnecessary when a circuit is evaluated on a batch of inputs, since the verifier can read the circuit once per batch \cite{bunz2026flock}.
\end{remark}

We track the following worst-case efficiency parameters of a (relativizing preprocessing) IOP:
\begin{itemize}[noitemsep]

  \item \defn{Proof length} $\IOPProofLength$: the total number of alphabet symbols in the proof strings.

  \item \defn{Proof query complexity} $\IOPQueryCount$: the total number of verifier queries to the proof strings.

  \item \defn{Round complexity} $\IOPRoundComplexity$: the number of back-and-forth rounds between the verifier and the prover. Queries to proof strings do not count as interaction rounds.

  \item \defn{Verifier oracle query complexity} $\IOPOracleQueryComplexity$: the number of verifier calls to the oracle $\Oracle$, when present. These calls are counted separately from proof queries.

  \item \defn{Verifier preprocessing time} $\IOPVerifierPreTime$: for an IOP with preprocessing (\cref{rem:iop-preprocessing}), the running time of $\IOPVerifierPre$, which bounds $|\IOPVerifierState|$.

  \item \defn{Verifier online time} $\IOPVerifierTime$: the running time of $\IOPVerifierPost$, with random access to $\IOPVerifierState$.

  \item \defn{Prover running time} $\IOPProverTime$: the honest prover's total running time.

\end{itemize}

\subsection{Error-correcting codes}\label{sec:codes}

An error-correcting code maps a message to a longer string, its codeword, so that any two codewords differ in many positions. A locally decodable code \cite{katz2000efficiency} lets a decoder recover any symbol of the message from a few queries to a (possibly corrupted) codeword, and a relaxed locally decodable code \cite{ben2004robust} also allows the decoder to refuse to answer.

\begin{definition}[Distance]\label{defn:distance}

  Let $\Alphabet$ be a finite alphabet and $\IOPProofLength$ a positive integer. The \defn{distance} $\EWDistance(\IOPProof,\IOPProof')$ between $\IOPProof,\IOPProof'\in\Alphabet^{\MathScriptStyle \IOPProofLength}$ is the fraction of positions in which they differ. The distance between $\IOPProof$ and a set $S\subseteq\Alphabet^{\MathScriptStyle \IOPProofLength}$ is
  \[
    \EWDistance(\IOPProof,S)\coloneq\min_{\MathScriptStyle \IOPProof'\in S}\EWDistance(\IOPProof,\IOPProof'),
  \]
  and $\EWDistance(\IOPProof,S)\coloneq1$ if $S$ is empty. For $\EWProximity\in(0,1)$, we say that $\IOPProof$ is \defn{$\EWProximity$-close} to $S$ if $\EWDistance(\IOPProof,S)\le\EWProximity$, and \defn{$\EWProximity$-far} from $S$ otherwise.

\end{definition}

\begin{definition}[Code]\label{defn:code}

  Let $\Alphabet$ be a finite alphabet and $\IOPProofLength$ a positive integer. A \defn{code} is a set $\Code\subseteq\Alphabet^{\MathScriptStyle \IOPProofLength}$, whose elements are called \defn{codewords}. Its \defn{relative distance} is the minimum distance (\cref{defn:distance}) between two distinct codewords,
  \[
    \EWDistance(\Code)\coloneq\min_{\MathScriptStyle \Codeword\ne\Codeword'\in\Code}\EWDistance(\Codeword,\Codeword').
  \]
  For a finite alphabet $\EWMessageAlphabet$ and an integer $\EWLength\ge0$, an \defn{encoding} is an injective map $\EWEncoding:\EWMessageAlphabet^{\MathScriptStyle \EWLength}\to\Alphabet^{\MathScriptStyle \IOPProofLength}$. Its image is a code, and the relative distance of $\EWEncoding$ is that of its image.

\end{definition}

We write $\PositionList=(i_{\MathScriptStyle 1},\ldots,i_{\MathScriptStyle |\PositionList|})\in[\EWLength]^{\MathScriptStyle *}$ for a list of \defn{positions} and $\RestrictTo{\Witness}{\PositionList}\coloneq(\Witness_{\MathScriptStyle i_1},\ldots,\Witness_{\MathScriptStyle i_{|\PositionList|}})$ for the symbols of $\Witness\in\EWMessageAlphabet^{\MathScriptStyle \EWLength}$ at these positions.

\begin{definition}[Relaxed local decoder]\label{defn:relaxed-local-decoder}

  Let $\EWEncoding:\EWMessageAlphabet^{\MathScriptStyle \EWLength}\to\Alphabet^{\MathScriptStyle \IOPProofLength}$ be an encoding. A \defn{relaxed local decoder} for $\EWEncoding$ with \defn{query complexity} $\DecoderQueryCount$ is a probabilistic algorithm $\RelaxedDecoder$ that receives a list $\PositionList$ of positions, has oracle access to a string $\Codeword\in\Alphabet^{\MathScriptStyle \IOPProofLength}$, reads at most $\DecoderQueryCount$ symbols of $\Codeword$, and outputs $\DecodedBits\in\EWMessageAlphabet^{\MathScriptStyle |\PositionList|}\cup\{\bot\}$. We write $\DecoderRun{\Codeword}$ for its output.

\end{definition}

\begin{definition}[Completeness]\label{definition:rld-completeness}

  A relaxed local decoder for $\EWEncoding$ has \defn{perfect completeness} if, for every $\Witness\in\EWMessageAlphabet^{\MathScriptStyle \EWLength}$ and every list $\PositionList$ of positions,
  \[
    \probability{
      \DecodedBits=\RestrictTo{\Witness}{\PositionList}
      \;\middle\vert\;
      \DecoderRun{\EWEncoding(\Witness)}
    }=1.
  \]

\end{definition}

\begin{definition}[Decoding error]\label{definition:rld-decoding-error}

  A relaxed local decoder for $\EWEncoding$ has \defn{decoding error} $\ExtensionError:(0,1)\times\bbN\to[0,1]$ if, for every \defn{proximity parameter} $\EWProximity\in(0,1)$, every list $\PositionList$ of positions, every $\Witness\in\EWMessageAlphabet^{\MathScriptStyle \EWLength}$, and every $\Malicious{\Codeword}\in\Alphabet^{\MathScriptStyle \IOPProofLength}$ that is $\EWProximity$-close to $\EWEncoding(\Witness)$ (\cref{defn:distance}),
  \[
    \probability{
      \DecodedBits\notin\{\RestrictTo{\Witness}{\PositionList},\bot\}
      \;\middle\vert\;
      \DecoderRun{\Malicious{\Codeword}}
    }\le\ExtensionError(\EWProximity,|\PositionList|).
  \]

\end{definition}

In this paper we use a more relaxed notion, interactive local decoding, in which a prover helps the decoder. This relaxation is what enables efficient proving and verification: the encoding of \cref{lem:ild-encoding} has linear length, and decoding a few positions costs the prover linear time and the decoder polylogarithmic time in the message length.

\begin{definition}[Interactive local decoder]\label{defn:iopp}

  Let $\EWMessageAlphabet\supseteq\bits$ be a finite alphabet and let $\EWEncoding:\EWMessageAlphabet^{\MathScriptStyle \EWLength}\to\Alphabet^{\MathScriptStyle \IOPProofLength}$ be an encoding (\cref{defn:code}). An \defn{interactive local decoder} for $\EWEncoding$ is an interactive protocol between a probabilistic polynomial-time \defn{decoder} $\ExternalVerifier$ and a possibly computationally unbounded prover $\TestProver$. The parties receive a common input $\PositionList$, a list of positions. The decoder has oracle access to a string $\IOPProof\in\Alphabet^{\MathScriptStyle \IOPProofLength}$, and the honest prover receives $\Witness\in\EWMessageAlphabet^{\MathScriptStyle \EWLength}$ with $\IOPProof=\EWEncoding(\Witness)$. The prover sends proof strings as in an IOP (\cref{defn:iop}), and at the end the decoder outputs $\DecodedBits\in\EWMessageAlphabet^{\MathScriptStyle |\PositionList|}\cup\{\bot\}$.

\end{definition}

We write $\EWTestRun{\TestProver(\Witness)}{\IOPProof}$ for the output of the decoder. We track the efficiency parameters of an interactive local decoder as for an IOP (\cref{section:iop}), with the queries to $\IOPProof$ counted as proof queries.

\begin{definition}[Completeness]\label{definition:iopp-completeness}

  An interactive local decoder for $\EWEncoding$ has \defn{perfect completeness} if, for every $\Witness\in\EWMessageAlphabet^{\MathScriptStyle \EWLength}$ and every list $\PositionList$ of positions, the honest prover satisfies
  \[
    \probability{
      \DecodedBits=\RestrictTo{\Witness}{\PositionList}
      \;\middle\vert\;
      \EWTestRun{\TestProver(\Witness)}{\EWEncoding(\Witness)}
    }=1.
  \]

\end{definition}

\begin{definition}[Decoding error]\label{definition:iopp-soundness}

  An interactive local decoder for $\EWEncoding$ has \defn{decoding error} $\ExtensionError:(0,1)\times\bbN\to[0,1]$ if, for every \defn{proximity parameter} $\EWProximity\in(0,1)$, every list $\PositionList$ of positions, every $\Witness\in\EWMessageAlphabet^{\MathScriptStyle \EWLength}$, every $\Malicious{\IOPProof}\in\Alphabet^{\MathScriptStyle \IOPProofLength}$ that is $\EWProximity$-close to $\EWEncoding(\Witness)$ (\cref{defn:distance}), and every possibly computationally unbounded cheating prover $\MaliciousTestProver$,
  \[
    \probability{
      \DecodedBits\notin\{\RestrictTo{\Witness}{\PositionList},\bot\}
      \;\middle\vert\;
      \EWTestRun{\MaliciousTestProver}{\Malicious{\IOPProof}}
    }\le\ExtensionError(\EWProximity,|\PositionList|).
  \]

\end{definition}

\begin{lemma}[General approach \cite{zeilberger2024basefold,arnon2025whir,novakovic2025ligerito}; linear-time construction \cite{bunz2026tensorswitch,spielman1996linear}]\label{lem:ild-encoding}

  Let $\EWLength\ge1$. There is a prime field $\Field$ of size at least a sufficiently large polynomial in $\EWLength$ and at most polynomial in $\EWLength$, and an injective encoding $\EWEncoding:\Field^{\MathScriptStyle \EWLength}\to\Field^{\MathScriptStyle \IOPProofLength}$ with $\IOPProofLength=\BigO{\EWLength}$, computable in $\BigO{\EWLength}$ field operations, with relative distance at least a constant $\EWDistanceBound>0$, such that for every constant proximity parameter $\EWProximity\in(0,\EWDistanceBound/4)$ there is an interactive local decoder for $\EWEncoding$ that, on a list $\PositionList$ of positions, has the following parameters, where the constants in the $O(\cdot)$ depend on $\EWProximity$.
  \begin{itemize}[noitemsep]
    \item \defn{Perfect completeness}.
    \item \defn{Decoding error}: $\ExtensionError(\EWProximity,|\PositionList|)\le1/12$.
    \item \defn{Round complexity}: $\BigO{\log\EWLength}$.
    \item \defn{Proof query complexity}: $\BigO{\log\EWLength}$.
    \item \defn{Proof length}: $\BigO{\EWLength}$.
    \item \defn{Prover time}: $\BigO{\EWLength+|\PositionList|}$.
    \item \defn{Verifier time}: $\BigO{(|\PositionList|+1)\log\EWLength}$.
  \end{itemize}
  The decoder's verifier has oracle access to a string of length $\EWLength^{\MathScriptStyle 1/2+o(1)}$ that depends only on $\EWLength$ and is computable in that time; in \cref{lem:ewiop-circuits} the verifier computes it in its preprocessing.

\end{lemma}

\begin{proof}[Proof sketch]
  The decoder follows the multilinear polynomial commitments of BaseFold, WHIR, and Ligerito \cite{zeilberger2024basefold,arnon2025whir,novakovic2025ligerito}, which check a claimed evaluation of the multilinear extension of a message against its encoding; TensorSwitch \cite{bunz2026tensorswitch} is the linear-time instantiation that we use. $\EWEncoding$ is the encoding of TensorSwitch, a tensor code of two linear-time encodable codes of constant rate and constant relative distance \cite{spielman1996linear}, so it has linear length, constant relative distance, and linear encoding time. TensorSwitch binds a string to a message by decoding its columns and then its rows, each within the unique decoding radius of the respective code; a string that is $\EWProximity$-close to $\EWEncoding(\Witness)$ with $\EWProximity<\EWDistanceBound/4$ has few enough undecodable columns for this to recover $\Witness$, and in this regime the hypotheses of TensorSwitch on the codes hold for every linear code. The decoder works as follows. The prover sends the claimed values $\DecodedBits'\in\Field^{\MathScriptStyle |\PositionList|}$ of the positions in $\PositionList$. The decoder picks a random $r\in\Field^{\MathScriptStyle |\PositionList|}$, and the parties run a sumcheck for the claim $\sum_{i}u_{\MathScriptStyle i}\Witness_{\MathScriptStyle i}=\sum_{k}r_{\MathScriptStyle k}\DecodedBits'_{\MathScriptStyle k}$, where $u\coloneq\sum_{k}r_{\MathScriptStyle k}e_{\MathScriptStyle i_k}$ has $|\PositionList|$ nonzero entries. This reduces the claim to one evaluation of the multilinear extension of $\Witness$ at a random point, which the parties check against $\IOPProof$ by running the remaining rounds of TensorSwitch's commitment phase and its opening protocol. The decoder outputs $\DecodedBits'$ if every check passes and $\bot$ otherwise. If some $\DecodedBits'_{\MathScriptStyle k}\ne\Witness_{\MathScriptStyle i_k}$, the combined claim is false except with probability $1/|\Field|$ over $r$, and then the sumcheck or the opening rejects except with error polynomially small in $|\Field|$ plus the error of the opening, which is $2^{\MathScriptStyle -\lambda}$ for $\BigO{\lambda}$ queries; $\lambda=4$ gives decoding error below $1/12$. The prover runs the sumcheck in $\BigO{\EWLength}$ field operations, since $u$ is sparse, and the opening in $\BigO{\EWLength}$. The verifier runs in $\BigO{\log\EWLength}$ for the sumcheck, $\BigO{|\PositionList|\log\EWLength}$ to evaluate the multilinear extension of $u$ at the random point, and $\BigO{\log\EWLength}$ for the opening, which reads a preprocessed string of length $\EWLength^{\MathScriptStyle 1/2+o(1)}$ that depends only on the code.
\end{proof}

\subsection{Encoded-witness IOPs}\label{sec:ewiop}
Here we define the notion of encoded-witness IOPs, which serve as a building block in our relativizing IOP in \cref{sec:relativized-iops}.
An encoded-witness IOP is an IOP in which the prover's first message is an encoding of its witness. Soundness is stated in terms of this first message \cite{ben2016interactive,ben2018fri}. In addition, the encoding is required to have an interactive local decoder (\cref{defn:iopp}).

\begin{definition}[Encoded-witness IOP]\label{defn:ewiop}

  Let $\Relation$ be a relation whose witnesses are strings $\Witness\in\bits^{\MathScriptStyle \EWLength}$, and let $\EWEncoding:\EWMessageAlphabet^{\MathScriptStyle \EWLength}\to\Alphabet^{\MathScriptStyle \IOPProofLength}$ be an encoding (\cref{defn:code}). An \defn{encoded-witness IOP} for $\Relation$ with encoding $\EWEncoding$ is an IOP with preprocessing $(\IOPProver,\IOPVerifier)$ for $\Relation$ (\cref{defn:iop,rem:iop-preprocessing}) in which the prover's first message is a proof string $\IOPProof_{\MathScriptStyle 1}\in\Alphabet^{\MathScriptStyle \IOPProofLength}$, and the honest prover sends $\IOPProof_{\MathScriptStyle 1}\coloneq\EWEncoding(\Witness)$, where the encoding $\EWEncoding$ has an interactive local decoder $(\TestProver,\ExternalVerifier)$ (\cref{defn:iopp}). We write
  \[
    \EWIOPRun{\IOPProver(\Witness)}
  \]
  where $\IOPProof_{\MathScriptStyle 1}$ denotes the first message of the prover and $\EWDecision$ denotes the decision bit of the verifier. The decoder has the completeness and decoding error of \cref{definition:iopp-completeness,definition:iopp-soundness}; the IOP has the completeness and soundness properties defined below.

\end{definition}

\begin{definition}[Completeness]\label{definition:ewiop-completeness}

  An encoded-witness IOP has \defn{perfect completeness} if, for every $(\Instance,\Witness)\in\Relation$, the honest prover satisfies
  \[
    \probability{
      \EWDecision=1
      \;\middle\vert\;
      \EWIOPRun{\IOPProver(\Witness)}
    }=1.
  \]

\end{definition}

\begin{definition}[Soundness]\label{definition:ewiop-soundness}

  For an instance $\Instance$, let $\EWValidCode\coloneq\{\EWEncoding(\Witness):(\Instance,\Witness)\in\Relation\}$ be the set of encodings of valid witnesses. An encoded-witness IOP has \defn{soundness error} $\IOPSoundnessError:(0,1)\to[0,1]$ if, for every \defn{proximity parameter} $\EWProximity\in(0,1)$, every instance $\Instance$, and every possibly computationally unbounded cheating prover $\Malicious{\IOPProver}$,
  \[
    \probability{
      \EWDecision=1\;\land\;
      \EWDistance\bigl(\IOPProof_{\MathScriptStyle 1},\EWValidCode\bigr)>\EWProximity
      \;\middle\vert\;
      \EWIOPRun{\Malicious{\IOPProver}}
    }\le\IOPSoundnessError(\EWProximity).
  \]

\end{definition}

\begin{remark}[Holographic proofs]\label{rem:holographic}
  Encoded-witness IOPs are similar to holographic PCPs \cite{babai1991checking}, PCPs of proximity \cite{ben2004robust}, and IOPs of proximity \cite{ben2016interactive,reingold2016constant}, in which the verifier has oracle access to an input and certifies that it is close to a language; here the input is the prover's first message and the language is the set of encodings of valid witnesses. However, these proof systems only guarantee proximity to such an encoding, whereas we also need to recover positions of the witness from the first message, which is what the interactive local decoder provides.
\end{remark}

\begin{lemma}[Encoded-witness IOP for circuits, adapted from \cite{chen2022hyperplonk,chiesa2020marlin,bunz2020dark,chiesa2026fiatshamir,bunz2026tensorswitch}]\label{lem:ewiop-circuits}

  Consider instances of $\CSATRelation$ (\cref{defn:csat-relation}) whose circuit has a public input of length $\InstanceSize$, a private input of length $\PrivateInputSize$, output length $\OutputSize$, and $\CircuitSize$ gates. Let $\EncodedWitnessLength\coloneq\CircuitSize+\InstanceSize+\PrivateInputSize+\OutputSize$, and let $\EWEncoding$ be the encoding with an interactive local decoder given by \cref{lem:ild-encoding} for $\EWLength\coloneq\EncodedWitnessLength$, applied to the witness $(\PrivateInput,\Transcript)$ padded with zeros to length $\EncodedWitnessLength$. Then, for every constant proximity parameter $\EWProximity\in(0,\EWDistanceBound/4)$, with $\EWDistanceBound$ as in \cref{lem:ild-encoding}, $\CSATRelation$ has an encoded-witness IOP $(\EWProver,\EWVerifier)$ with encoding $\EWEncoding$ with the following parameters, where $\SoftO{\cdot}$ hides factors polylogarithmic in $\EncodedWitnessLength$ and the constants in the $O(\cdot)$ depend on $\EWProximity$.
  \begin{itemize}[noitemsep]
    \item \defn{Perfect completeness}.
    \item \defn{Soundness error}: $\IOPSoundnessError(\EWProximity)\le1/12$.
    \item \defn{Round complexity}: $\BigO{\log\EncodedWitnessLength}$.
    \item \defn{Proof query complexity}: $\BigO{\log\EncodedWitnessLength}$.
    \item \defn{Proof length}: $\BigO{\EncodedWitnessLength}$.
    \item \defn{Prover time}: $\BigO{\EncodedWitnessLength}$.
    \item \defn{Verifier preprocessing time}: $\SoftO{\EncodedWitnessLength}$.
    \item \defn{Verifier online time}: $\SoftO{\InstanceSize+\OutputSize}$.
  \end{itemize}

\end{lemma}

\begin{proof}[Proof sketch]
  The IOP is the standard compilation of a polynomial IOP into an IOP \cite{chiesa2020marlin,bunz2020dark,chiesa2026fiatshamir}, applied to the polynomial IOP for circuit satisfiability of HyperPlonk \cite{chen2022hyperplonk}, with polynomials sent as codewords of the encoding $\EWEncoding$ of \cref{lem:ild-encoding} and evaluation claims checked by the proximity test of TensorSwitch \cite{bunz2026tensorswitch}. The prover's first message is $\EWEncoding(\PrivateInput,\Transcript)$, the encoding of the witness. The verifier's preprocessing encodes the polynomials that describe the circuit and computes the string of \cref{lem:ild-encoding}, which is its $\SoftO{\EncodedWitnessLength}$ cost; its online work is that of the polynomial IOP's verifier, which reads only the public input and output, plus a polylogarithmic number of queries. Whenever the polynomial IOP's verifier queries a polynomial at a point, the prover answers with the evaluation, and the two run the proximity test on the encoding and the claimed evaluation. For soundness, since $\EWProximity$ is below the unique decoding radius, every queried string is $\EWProximity$-close to at most one codeword, and the proximity test ensures, except with its error, that every answer is an evaluation of the polynomial that this codeword encodes. The soundness of the polynomial IOP then applies to the polynomial encoded by the first message, so if the verifier accepts with probability above $1/12$, the first message is $\EWProximity$-close to the encoding of a valid witness. The prover runs in $\BigO{\EncodedWitnessLength}$ field operations, since the prover of HyperPlonk, the encoding, and the prover of the proximity test all do. The constants in the $O(\cdot)$ depend on $\EWProximity$ through the proximity test's query complexity.
\end{proof}

\section{Relativizing Interactive Oracle Proofs}\label{sec:relativized-iops}\label{sec:encoded-witness-iops}
In this section we construct a relativizing IOP (\cref{defn:relativized-iop}) for the oracle-circuit-evaluation relation $\CvalRelation$ of \cref{defn:cval-relation}, that is, for proving that $\OracleCircuit(\CircuitInput)=\OutputValue$. 
\begin{thm}[Relativizing IOP for robust oracle circuits]
\label{ewiop-to-reliop}

  Fix an oracle $\Oracle:\bits^{\MathScriptStyle *}\to\bits$ and $0<\eps\le1$. Let $\OracleCircuit$ be an oracle circuit with an input of length $\InstanceSize$, an output of length $\OutputSize$, and $\CircuitSize$ gates, and let $\EncodedWitnessLength\coloneq\CircuitSize+\InstanceSize+\OutputSize$. Suppose that the oracle queries of $\OracleCircuit$ have length polylogarithmic in $\EncodedWitnessLength$. Then $\CvalRelation$, for instances $(\langle\OracleCircuit\rangle,\CircuitInput,\OutputValue)$ with this circuit such that $\OracleCircuit$ is $\eps$-robust with respect to $\CircuitInput$, has a relativizing IOP with preprocessing (\cref{rem:iop-preprocessing}) over a prime field of size polynomial in $\EncodedWitnessLength$ with the following properties, where $\SoftO{\cdot}$ hides factors polylogarithmic in $\EncodedWitnessLength$, running times count field operations, and proof lengths count field elements.

  \begin{itemize}[noitemsep]
    \item \defn{Perfect completeness}.
    \item \defn{Soundness error}: at most $1/3$.
    \item \defn{Round complexity}: $\BigO{\log\EncodedWitnessLength}$.
    \item \defn{Proof query complexity}: $\BigO{\log\EncodedWitnessLength}$.
    \item \defn{Verifier oracle queries}: $\BigO{1/\eps}$.
    \item \defn{Proof length}: $\BigO{\EncodedWitnessLength}$.
    \item \defn{Prover time}: $\BigO{\EncodedWitnessLength}+\SoftO{1/\eps}$.
    \item \defn{Verifier preprocessing time}: $\SoftO{\EncodedWitnessLength}$.
    \item \defn{Verifier online time}: $\SoftO{\InstanceSize+\OutputSize+1/\eps}$.
  \end{itemize}

\end{thm}

Recall that we want a relativizing IOP for $\CvalRelation$, that is, for the claim that $\OracleCircuit(\CircuitInput)=\OutputValue$, in which the verifier makes only a few queries to $\Oracle$. Without oracle gates this would be ordinary circuit satisfiability, which standard IOPs handle; the difficulty is the oracle gates, which the verifier cannot evaluate itself without making $\OracleGateCount$ oracle queries. Intuitively, we can ask the prover to supply the answer to every oracle query. This turns $\OracleCircuit$ into its de-oraclized circuit (\cref{defn:deoraclized}), an ordinary circuit whose evaluation an IOP can verify, and leaves the verifier with the task of checking that the supplied answers agree with $\Oracle$. Checking all of them would again take $\OracleGateCount$ oracle queries, but by robustness a wrong output can only be reached by lying about more than an $\eps$-fraction of the answers, so a few random spot checks suffice. Concretely, the construction combines the encoded-witness IOP for circuits of \cref{lem:ewiop-circuits} with random sampling of oracle gates. The prover encodes the evaluation of the de-oraclized circuit of $\OracleCircuit$ (\cref{defn:deoraclized}) on input $\CircuitInput$ and the oracle answers in its transcript, and proves that the de-oraclized circuit outputs $\OutputValue$. The verifier then samples a few oracle gates, decodes the query and the recorded answer at each of them from the encoded witness, and checks that each answer agrees with the oracle. By robustness (\cref{defn:robustness}), a wrong output forces more than an $\eps$-fraction of the recorded answers to be wrong, so $\BigO{1/\eps}$ samples suffice. \Cref{sec:oracle-iop-construction} gives the protocol and \cref{sec:oracle-iop-analysis} proves the theorem.

\subsection{Construction}\label{sec:oracle-iop-construction}

We describe the protocol $(\IOPProver,\IOPVerifier)$ for $\CvalRelation$. Fix $\Oracle$, $\eps$, and a field $\Field$ as in \cref{ewiop-to-reliop}, an instance $\Instance=(\langle\OracleCircuit\rangle,\CircuitInput,\OutputValue)$ with $\InstancePre=\langle\OracleCircuit\rangle$ and $\InstancePost=(\CircuitInput,\OutputValue)$, and the witness $\Transcript=\TranscriptFunction_{\MathScriptStyle \OracleCircuit}(\CircuitInput)$. Let $\Circuit$ be the de-oraclized circuit of $\OracleCircuit$ (\cref{defn:deoraclized}), let $\PrivateInput\in\bits^{\MathScriptStyle \OracleGateCount}$ be the values of $\Transcript$ at the oracle gates, and let
\[
  \CandidateTable\coloneq\PrivateInput\mathbin{\|}\TranscriptFunction_{\MathScriptStyle \Circuit}(\CircuitInput,\PrivateInput)
\]
be the witness of $\CSATRelation$ (\cref{defn:csat-relation}) for the instance $(\langle\Circuit\rangle,\CircuitInput,\OutputValue)$. The answer at oracle gate $j$ is the bit $\PrivateInput_{\MathScriptStyle j}$ of $\CandidateTable$, and each bit of its query is a bit of $\CircuitInput$ or of $\CandidateTable$, at a position determined by $\langle\OracleCircuit\rangle$. Let $(\EWProver,\EWVerifier)$ be the encoded-witness IOP of \cref{lem:ewiop-circuits} for a constant proximity parameter $\EWProximity\in(0,\EWDistanceBound/4)$, let $\EWEncoding$ be its encoding, and let $(\TestProver,\ExternalVerifier)$ be the interactive local decoder of $\EWEncoding$ for proximity parameter $\EWProximity$ (\cref{lem:ild-encoding}). The protocol is as follows.

\begin{enumerate}
  \item \defn{IOP}: $\IOPProver$ and $\IOPVerifier$ run $(\EWProver,\EWVerifier)$ on $(\langle\Circuit\rangle,\CircuitInput,\OutputValue)$ with witness $\CandidateTable$. This gives the first message $\IOPProof_{\MathScriptStyle 1}=\EWEncoding(\CandidateTable)$ and the decision bit $\EWDecision$.

  \item \defn{Decoding}: $\IOPVerifier$ samples a uniformly random set $\{j_{\MathScriptStyle 1},\ldots,j_{\MathScriptStyle \SampleCount}\}$ of $\SampleCount\coloneq\min\{\OracleGateCount,\lceil(\ln6)/\eps\rceil\}$ oracle gates and lets $\PositionList$ be the positions of their answers and query bits. Then $\IOPProver$ and $\IOPVerifier$ run $(\TestProver,\ExternalVerifier)$ on $\IOPProof_{\MathScriptStyle 1}$ and $\PositionList$, which gives $\DecodedBits\in\bits^{\MathScriptStyle |\PositionList|}\cup\{\bot\}$. Unless $\DecodedBits=\bot$, this yields for each $k\in[\SampleCount]$ a decoded answer $\Malicious{\OracleAnswer}_{\MathScriptStyle k}$ at the $j_{\MathScriptStyle k}$-th oracle gate and, together with the bits of $\CircuitInput$, a decoded query $\Malicious{\OracleQuery}_{\MathScriptStyle k}$.

  \item \defn{Decision}: $\IOPVerifier$ accepts if and only if $\EWDecision=1$, $\DecodedBits\ne\bot$, and $\Oracle(\Malicious{\OracleQuery}_{\MathScriptStyle k})=\Malicious{\OracleAnswer}_{\MathScriptStyle k}$ for every $k\in[\SampleCount]$.
\end{enumerate}

\subsection{Proof of \texorpdfstring{\cref{ewiop-to-reliop}}{the main theorem}}\label{sec:oracle-iop-analysis}

  \paragraph{Completeness.} Suppose that $\OracleCircuit(\CircuitInput)=\OutputValue$. Then $\PrivateInput$ holds the answers of $\Oracle$ in the evaluation of $\OracleCircuit$ on $\CircuitInput$, so $\Circuit(\CircuitInput,\PrivateInput)=\OutputValue$ and $(\PrivateInput,\TranscriptFunction_{\MathScriptStyle \Circuit}(\CircuitInput,\PrivateInput))$ is a valid witness. Hence $\EWDecision=1$ with probability $1$ by the completeness of the encoded-witness IOP, and $\DecodedBits=\RestrictTo{\CandidateTable}{\PositionList}$ with probability $1$ by the completeness of the decoder (\cref{lem:ild-encoding}). The decoded bits are bits of $\CandidateTable$, so $\Malicious{\OracleQuery}_{\MathScriptStyle k}$ is the query at the $j_{\MathScriptStyle k}$-th oracle gate and $\Malicious{\OracleAnswer}_{\MathScriptStyle k}=\Oracle(\Malicious{\OracleQuery}_{\MathScriptStyle k})$.

  \paragraph{Soundness.} Consider an instance $\Instance=(\langle\OracleCircuit\rangle,\CircuitInput,\OutputValue)$ with $\OracleCircuit(\CircuitInput)\ne\OutputValue$, and let $\Malicious{\IOPProver}$ be a cheating prover for our IOP. Since $\EWProximity<\EWDistanceBound/4$ and $\EWEncoding$ has relative distance at least $\EWDistanceBound$ (\cref{lem:ild-encoding}), the first message $\IOPProof_{\MathScriptStyle 1}$ of $\Malicious{\IOPProver}$ is $\EWProximity$-close to the encoding of at most one string. Let $(\PrivateInput^{\MathScriptStyle *},\Transcript^{\MathScriptStyle *})$ be this string if it is a valid witness of $\CSATRelation$ for $(\langle\Circuit\rangle,\CircuitInput,\OutputValue)$, and $\bot$ otherwise; it is determined by $\IOPProof_{\MathScriptStyle 1}$. We consider three events:
  \begin{itemize}[noitemsep]
    \item $E_{\MathScriptStyle 1}$: $(\PrivateInput^{\MathScriptStyle *},\Transcript^{\MathScriptStyle *})=\bot$ and $\EWDecision=1$;
    \item $E_{\MathScriptStyle 2}$: $(\PrivateInput^{\MathScriptStyle *},\Transcript^{\MathScriptStyle *})\ne\bot$ and $\DecodedBits\notin\{\RestrictTo{(\PrivateInput^{\MathScriptStyle *},\Transcript^{\MathScriptStyle *})}{\PositionList},\bot\}$;
    \item $E_{\MathScriptStyle 3}$: $(\PrivateInput^{\MathScriptStyle *},\Transcript^{\MathScriptStyle *})\ne\bot$, $\DecodedBits=\RestrictTo{(\PrivateInput^{\MathScriptStyle *},\Transcript^{\MathScriptStyle *})}{\PositionList}$, and $\Oracle(\Malicious{\OracleQuery}_{\MathScriptStyle k})=\Malicious{\OracleAnswer}_{\MathScriptStyle k}$ for every $k\in[\SampleCount]$.
  \end{itemize}
  If $\IOPVerifier$ accepts, then $\EWDecision=1$, $\DecodedBits\ne\bot$, and every oracle check passes, so one of $E_{\MathScriptStyle 1}$, $E_{\MathScriptStyle 2}$, $E_{\MathScriptStyle 3}$ occurs. Suppose toward a contradiction that
  \[
    \probability{\langle\Malicious{\IOPProver},\IOPVerifier\rangle(\Instance)=1}>\frac{1}{3}.
  \]
  Then, by the union bound, $\probability{E_{\MathScriptStyle 1}}>1/12$, $\probability{E_{\MathScriptStyle 2}}>1/12$, or $\probability{E_{\MathScriptStyle 3}}>1/6$, and we show that each case is impossible.
  \begin{itemize}[noitemsep]
    \item Suppose $\probability{E_{\MathScriptStyle 1}}>1/12$. Let $\MaliciousEWProver$ be the cheating prover for the encoded-witness IOP $(\EWProver,\EWVerifier)$ that runs $\Malicious{\IOPProver}$ in the IOP step and then stops; the first message and the decision bit in this run are distributed as $\IOPProof_{\MathScriptStyle 1}$ and $\EWDecision$ in our protocol. When $(\PrivateInput^{\MathScriptStyle *},\Transcript^{\MathScriptStyle *})=\bot$, the first message $\IOPProof_{\MathScriptStyle 1}$ is $\EWProximity$-far from $\EWValidCode$, the set of encodings of valid witnesses of $\CSATRelation$ for $(\langle\Circuit\rangle,\CircuitInput,\OutputValue)$, because it is $\EWProximity$-close to the encoding of at most one string and that string is not a valid witness. Therefore
    \[
      \probability{
        \EWDecision=1\;\land\;
        \EWDistance(\IOPProof_{\MathScriptStyle 1},\EWValidCode)>\EWProximity
        \;\middle\vert\;
        (\IOPProof_{\MathScriptStyle 1},\EWDecision)\leftarrow\langle\MaliciousEWProver,\EWVerifier\rangle(\langle\Circuit\rangle,\CircuitInput,\OutputValue)
      }
      \ge\probability{E_{\MathScriptStyle 1}}>\frac{1}{12}\ge\IOPSoundnessError(\EWProximity),
    \]
    contradicting the soundness of \cref{lem:ewiop-circuits} (\cref{definition:ewiop-soundness}).
    \item Suppose $\probability{E_{\MathScriptStyle 2}}>1/12$. By averaging, there is a transcript of the IOP step together with a choice of the sampled gates, which fix $\IOPProof_{\MathScriptStyle 1}$, $(\PrivateInput^{\MathScriptStyle *},\Transcript^{\MathScriptStyle *})\ne\bot$, and $\PositionList$, conditioned on which $E_{\MathScriptStyle 2}$ occurs with probability greater than $1/12$. Let $\MaliciousTestProver$ be the cheating prover for the decoder $(\TestProver,\ExternalVerifier)$ that has this transcript and these gates hardwired and continues to run $\Malicious{\IOPProver}$ from the decoding step; the output of $\ExternalVerifier$ in this run is distributed as $\DecodedBits$ in our protocol under the same conditioning. Since $\IOPProof_{\MathScriptStyle 1}$ is $\EWProximity$-close to $\EWEncoding(\PrivateInput^{\MathScriptStyle *},\Transcript^{\MathScriptStyle *})$, we get
    \[
      \probability{
        \DecodedBits\notin\{\RestrictTo{(\PrivateInput^{\MathScriptStyle *},\Transcript^{\MathScriptStyle *})}{\PositionList},\bot\}
        \;\middle\vert\;
        \EWTestRun{\MaliciousTestProver}{\IOPProof_{\MathScriptStyle 1}}
      }
      >\frac{1}{12}\ge\ExtensionError(\EWProximity,|\PositionList|),
    \]
    contradicting the decoding error of \cref{lem:ild-encoding} (\cref{definition:iopp-soundness}).
    \item Suppose $\probability{E_{\MathScriptStyle 3}}>1/6$. When $(\PrivateInput^{\MathScriptStyle *},\Transcript^{\MathScriptStyle *})\ne\bot$, it is a valid witness, so $\Circuit(\CircuitInput,\PrivateInput^{\MathScriptStyle *})=\OutputValue\ne\OracleCircuit(\CircuitInput)$, and by robustness (\cref{defn:robustness}) $\Oracle(\OracleQuery_{\MathScriptStyle j}(\PrivateInput^{\MathScriptStyle *}))\ne\PrivateInput^{\MathScriptStyle *}_{\MathScriptStyle j}$ for more than $\eps\OracleGateCount$ indices $j\in[\OracleGateCount]$, with $\OracleQuery_{\MathScriptStyle j}(\cdot)$ as in that definition. On $E_{\MathScriptStyle 3}$ the decoded bits are those of $(\PrivateInput^{\MathScriptStyle *},\Transcript^{\MathScriptStyle *})$, so the $k$-th decoded pair is $(\OracleQuery_{\MathScriptStyle j_{\MathScriptStyle k}}(\PrivateInput^{\MathScriptStyle *}),\PrivateInput^{\MathScriptStyle *}_{\MathScriptStyle j_{\MathScriptStyle k}})$, and every oracle check passes only if no sampled gate is one of these indices. The sampled set is uniformly random and independent of $(\PrivateInput^{\MathScriptStyle *},\Transcript^{\MathScriptStyle *})$, which is determined by $\IOPProof_{\MathScriptStyle 1}$, sent before the sampling. Hence $\probability{E_{\MathScriptStyle 3}}\le(1-\eps)^{\MathScriptStyle \SampleCount}\le e^{\MathScriptStyle -\eps\SampleCount}\le1/6$ if $\SampleCount=\lceil(\ln6)/\eps\rceil$, and $\probability{E_{\MathScriptStyle 3}}=0$ if $\SampleCount=\OracleGateCount$, a contradiction.
  \end{itemize}
  Hence $\IOPVerifier$ accepts with probability at most $1/3$. If $\OracleGateCount=0$, then $\Circuit(\CircuitInput)=\OracleCircuit(\CircuitInput)\ne\OutputValue$, so there is no valid witness and only $E_{\MathScriptStyle 1}$ can occur.

  \paragraph{Efficiency.} The circuit $\Circuit$ has $\OracleGateCount\le\CircuitSize$ private input bits and $\CircuitSize-\OracleGateCount$ gates, so \cref{lem:ewiop-circuits} applies to it with $\EncodedWitnessLength=\CircuitSize+\InstanceSize+\OutputSize$ as in the statement of the theorem, and the list $\PositionList$ has at most $\SampleCount\cdot\OraclePairSize\le\lceil(\ln6)/\eps\rceil\cdot\OraclePairSize=\SoftO{1/\eps}$ positions, where $\OraclePairSize$ is one plus the maximum fan-in of an oracle gate, which is polylogarithmic by hypothesis. The round complexity, proof query complexity, proof length, and prover time are the sums of those of the IOP in \cref{lem:ewiop-circuits} and of the decoder in \cref{lem:ild-encoding}; in particular the prover time is $\BigO{\EncodedWitnessLength}+\BigO{|\PositionList|}=\BigO{\EncodedWitnessLength}+\SoftO{1/\eps}$. The verifier preprocessing time is that of the encoded-witness IOP plus the time $\SoftO{\CircuitSize}$ to compute $\langle\Circuit\rangle$ and the positions from $\langle\OracleCircuit\rangle$. The verifier makes $\SampleCount=\BigO{1/\eps}$ oracle queries, and its online time is that of the IOP plus that of the decoder plus $\SoftO{|\PositionList|}$ to read the positions and form the queries, for a total of $\SoftO{\InstanceSize+\OutputSize+1/\eps}$.

\section{Relativizing Rewindable Interactive Proofs}\label{sec:rewindable}\label{sec:robust}
In this section we formally define relativizing rewindable interactive proofs (\cref{sec:rip-definition}), show how to turn any relativizing IOP into a relativizing RIP (\cref{sec:iop-to-rip}), and define the efficiency that we aim for and prove our main theorem (\cref{sec:robust-rip-proof}); in \cref{sec:rip-vs-mip} we show that RIPs are as powerful as MIPs.

\subsection{Definition}\label{sec:rip-definition}

As mentioned in \cref{sec:intro}, when the prover is an AI model, the verifier can save a snapshot of the model's state and later restore it. We capture this by letting the verifier query the prover on any history of its choice. Since the prover may be assumed deterministic (\cref{rem:deterministic-prover}), it is determined by its \defn{next-message function}, which maps the common input $\Instance$ and a sequence $(\VerifierMessage_{\MathScriptStyle 1},\ldots,\VerifierMessage_{\MathScriptStyle j})$ of verifier messages to the prover's reply after receiving these messages in this order; querying it on a history is the same as restoring the snapshot taken after that history and sending the last message.

\begin{definition}[Rewindable interactive proof]\label{defn:rewindable-ip}

  A \defn{rewindable interactive proof} (RIP) for a relation $\Relation$ is a pair $(\pro,\ver)$ of a deterministic prover, given by its next-message function $\pro(\Instance,\cdot)$, and a probabilistic polynomial-time verifier $\ver$ with oracle access to $\pro(\Instance,\cdot)$, which it may query adaptively on any sequence of messages. The parties receive a common input $\Instance$, and the honest prover additionally receives a witness $\Witness$ with $(\Instance,\Witness)\in\Relation$. At the end, the verifier accepts or rejects.

  The protocol satisfies the completeness and soundness properties defined below.

\end{definition}

We write $\ver^{\pro(\Instance,\cdot)}(\Instance)=1$ when the verifier accepts after interacting with the prover on common input $\Instance$. The probabilities below are over the verifier's random bits.

\begin{definition}[Completeness]\label{definition:rip-completeness}

  An RIP for $\Relation$ has \defn{perfect completeness} if, for every $(\Instance,\Witness)\in\Relation$, the honest prover satisfies
  \[
    \probability{
      \ver^{\pro(\Instance,\Witness,\cdot)}(\Instance)=1
    }=1.
  \]

\end{definition}

\begin{definition}[Soundness]\label{definition:rip-soundness}

  An RIP for $\Relation$ has \defn{soundness error} $\RIPSoundnessError$ if, for every $\Instance\notin\Language$ and every function $\Malicious{\pro}$ from sequences of messages to replies,
  \[
    \probability{
      \ver^{\Malicious{\pro}}(\Instance)=1
    }\le\RIPSoundnessError(|\Instance|).
  \]

\end{definition}

\begin{definition}[Relativizing RIP]\label{defn:relativized-rip}

  Fix an oracle $\Oracle$ and a relativized relation $\OracleRelation$ as in \cref{defn:relativized-relation}. A \defn{relativizing RIP} for $\OracleRelation$ is an RIP for $\OracleRelation$ in which the verifier and the prover, including a cheating prover, have black-box access to the same oracle $\Oracle$. The protocol \defn{relativizes} if its completeness, soundness, and efficiency guarantees hold for $\OracleRelation$ for every oracle $\Oracle$.

\end{definition}

\begin{remark}[Preprocessing]\label{rem:rip-preprocessing}
  As in \cref{rem:iop-preprocessing}, an RIP \defn{with preprocessing} for instances of the form $\Instance=(\InstancePre,\InstancePost)$ has a verifier that is a pair $\ver=(\VerifierPre,\VerifierPost)$. First $\VerifierPre$ receives $\InstancePre$ and outputs a state $\VerifierState$. Then $\VerifierPost$ receives $\VerifierState$ and $\InstancePost$ and queries the prover. Completeness and soundness are as in \cref{definition:rip-completeness,definition:rip-soundness}.
\end{remark}

\begin{remark}[Deterministic prover]\label{rem:deterministic-prover}
  Assuming a deterministic prover loses no generality. For completeness, the honest prover's random tape can be fixed and included in its state. For soundness, a randomized cheating prover is a distribution over deterministic next-message functions, so its acceptance probability is at most that of the best of them, provided that a snapshot includes the prover's random tape.
\end{remark}

We consider the following efficiency parameters of a (relativizing preprocessing) RIP, where each query to $\pro(\Instance,\cdot)$ counts as one message from the verifier and one from the prover:
\begin{itemize}[noitemsep]

  \item \defn{Round complexity} $\RIPRoundComplexity$: the number of back-and-forth rounds between the verifier and the prover. Queries made in parallel count as one round.

  \item \defn{Prover query complexity} $\RIPQueryCount$: the number of verifier queries to $\pro(\Instance,\cdot)$.

  \item \defn{Communication complexity} $\RIPCommunicationComplexity$: the total number of bits in the verifier's queries and the prover's answers.

  \item \defn{Verifier oracle query complexity} $\RIPOracleQueryComplexity$: the number of verifier calls to an external oracle $\Oracle$, when present.

  \item \defn{Verifier preprocessing time} $\RIPVerifierPreTime$: for an RIP with preprocessing (\cref{rem:rip-preprocessing}), the running time of $\VerifierPre$, which bounds $|\VerifierState|$.

  \item \defn{Verifier online time} $\RIPVerifierTime$: the running time of $\VerifierPost$, with random access to $\VerifierState$.

  \item \defn{Prover running time} $\RIPProverTime$: the honest prover's total time to answer all queries.

\end{itemize}
Saving and restoring a snapshot is not charged; the number of restores is at most the prover query complexity.

\subsection{From IOPs to RIPs}\label{sec:iop-to-rip}

The compilation is inspired by the standard one from an IOP to a two-prover protocol \cite{mip1988,fortnow1994power}, as applied to IOPs in \cite[Appendix~A]{ben2016interactive}, in which one prover answers the verifier's proof queries and a second, non-communicating prover answers one of them at random to bind the first to fixed proof strings, at the cost of dividing the rejection probability by the query count. With a rewindable prover the verifier binds the prover itself: it obtains each symbol of a proof string by restoring the prover to its state when that string was fixed, so the prover cannot answer adaptively and the compilation loses nothing in soundness.

\begin{lemma}[IOP-to-RIP transformation]\label{rel-iop-to-rip}

  Let $(\IOPProver,\IOPVerifier)$ be a relativizing IOP (\cref{defn:relativized-iop}) for a relativized relation $\OracleRelation$ with perfect completeness, soundness error $\IOPSoundnessError$, round complexity $\IOPRoundComplexity$, and proof query complexity $\IOPQueryCount$. Then $(\IOPProver,\IOPVerifier)$ can be transformed into a relativizing rewindable interactive proof $(\pro,\ver)$ (\cref{defn:relativized-rip}) for $\OracleRelation$ with the following properties.
  \begin{itemize}[noitemsep]
    \item \defn{Perfect completeness}.
    \item \defn{Soundness error}: at most $\IOPSoundnessError$.
    \item \defn{Round complexity}: $\BigO{\IOPRoundComplexity+\IOPQueryCount}$.
    \item \defn{Verifier oracle query complexity}: the same as that of $\IOPVerifier$.
  \end{itemize}
  If $(\IOPProver,\IOPVerifier)$ has preprocessing (\cref{rem:iop-preprocessing}), then so does $(\pro,\ver)$ (\cref{rem:rip-preprocessing}), with the same verifier preprocessing. The verifier $\ver$ runs $\IOPVerifier$, and the honest prover runs $\IOPProver$ once, with additional work only to process the query addresses and replies.

\end{lemma}

\begin{proof}

  \paragraph{The protocol.} The parties receive $\Instance$, and the honest prover also receives $\Witness$. Query addresses are tagged so that a history ending in a query is distinguishable from one ending in a verifier message.
  \begin{itemize}[noitemsep]
    \item \defn{Prover $\pro$}: runs $\IOPProver$ on $(\Instance,\Witness)$. On receiving a verifier message $\VerifierMessage$, it forwards $\VerifierMessage$ to $\IOPProver$, which computes its next proof string, and replies with the empty message. On receiving a query $\ProofQuery$ after $j-1$ verifier messages, it replies with the symbol $\IOPProof_{\MathScriptStyle j}(\ProofQuery)$ of the $j$-th proof string.
    \item \defn{Verifier $\ver$}: runs $\IOPVerifier$ on $\Instance$ and makes $\IOPVerifier$'s oracle queries to $\Oracle$ itself. Whenever $\IOPVerifier$ wishes to read position $\ProofQuery$ of the $j$-th proof string, $\ver$ queries $\pro$ on the history $(\VerifierMessage_{\MathScriptStyle 1},\ldots,\VerifierMessage_{\MathScriptStyle j-1},\ProofQuery)$, where $\VerifierMessage_{\MathScriptStyle 1},\ldots,\VerifierMessage_{\MathScriptStyle j-1}$ are the messages that $\IOPVerifier$ sent before the $j$-th proof string, and hands the reply to $\IOPVerifier$ as that symbol. Queries that $\IOPVerifier$ makes at the same time are made in parallel, and a message that $\IOPVerifier$ reads in full is obtained by querying all of its positions. $\ver$ accepts if and only if $\IOPVerifier$ accepts.
  \end{itemize}
  In the snapshot view of \cref{defn:rewindable-ip}, querying $\pro$ on $(\VerifierMessage_{\MathScriptStyle 1},\ldots,\VerifierMessage_{\MathScriptStyle j-1},\ProofQuery)$ restores the prover to its state just after it fixed the $j$-th proof string and asks for one symbol of it, which is why the prover cannot answer adaptively.

  \paragraph{Completeness.} The honest prover's replies are the symbols of $\IOPProver$'s proof strings, so the simulated $\IOPVerifier$ sees exactly the honest IOP interaction and accepts with probability $1$.

  \paragraph{Soundness.} Consider an instance $\Instance\notin\GetLanguage{\OracleRelation}$ and a cheating prover $\Malicious{\pro}$, which we may assume to be deterministic (\cref{rem:deterministic-prover}). Define a cheating IOP prover $\Malicious{\IOPProver}$ that, having received the verifier messages $\VerifierMessage_{\MathScriptStyle 1},\ldots,\VerifierMessage_{\MathScriptStyle j-1}$, sends the proof string
  \[
    \FixedProofString_{\MathScriptStyle j}(\ProofQuery)
      \coloneq
      \Malicious{\pro}(\VerifierMessage_{\MathScriptStyle 1},\ldots,\VerifierMessage_{\MathScriptStyle j-1},\ProofQuery).
  \]
  This is a valid IOP prover strategy, since $\FixedProofString_{\MathScriptStyle j}$ depends only on messages available when string $j$ is sent. Every symbol that $\ver$ obtains from $\Malicious{\pro}$ is the corresponding symbol of $\FixedProofString_{\MathScriptStyle j}$, so the view of the simulated $\IOPVerifier$ against $\Malicious{\pro}$ is exactly its view against $\Malicious{\IOPProver}$, and $\ver$ accepts with probability at most $\IOPSoundnessError$.

  \paragraph{Efficiency.} The simulation uses the rounds of the IOP, with the queries of each round made in parallel, plus at most one round per query for queries that $\IOPVerifier$ makes adaptively, for $\BigO{\IOPRoundComplexity+\IOPQueryCount}$ rounds in total. The verifier's oracle queries are those of $\IOPVerifier$. The verifier's work beyond $\IOPVerifier$'s is to form the query addresses and copy the message prefixes, and the honest prover's work beyond $\IOPProver$'s is to restore snapshots and read symbols. Preprocessing is preserved since $\ver$ preprocesses exactly as $\IOPVerifier$ does, and the argument applies verbatim when all parties have access to an oracle $\Oracle$.

\end{proof}

\subsection{RIPs with linear-time provers and succinct verifiers}\label{sec:robust-rip-proof}\label{sec:succinct-rip}

In our setting, the honest prover should take not much more time than evaluating the circuit, and the verifier should do substantially less work than evaluating the circuit itself. To verify the evaluation of an oracle circuit $\OracleCircuit$, we use an RIP with preprocessing for the relation $\CvalRelation$ of \cref{defn:cval-relation}: the parties receive the instance $\Instance=(\langle\OracleCircuit\rangle,\CircuitInput,\OutputValue)$, where $\InstancePre=\langle\OracleCircuit\rangle$ and $\InstancePost=(\CircuitInput,\OutputValue)$, the honest prover receives the transcript $\Transcript$, and the claim is that $\OracleCircuit(\CircuitInput)=\OutputValue$.

We now prove our main theorem by applying the transformation of \cref{sec:iop-to-rip} to the relativizing IOP of \cref{sec:relativized-iops}.

\begin{thm}\label{thm:robust-rip-parameters}

  Fix $\Oracle:\bits^{\MathScriptStyle *}\to\bits$ and $0<\eps\le1$, and let $\OracleCircuit$ be an oracle circuit with an input of length $\InstanceSize$, an output of length $\OutputSize$, and $\CircuitSize$ gates, and let $\EncodedWitnessLength\coloneq\CircuitSize+\InstanceSize+\OutputSize$ as in \cref{ewiop-to-reliop}. Suppose that the oracle queries of $\OracleCircuit$ have length polylogarithmic in $\EncodedWitnessLength$. For every $\SoundnessParameter\ge1$, $\CvalRelation$, for instances $(\langle\OracleCircuit\rangle,\CircuitInput,\OutputValue)$ with this circuit such that $\OracleCircuit$ is $\eps$-robust with respect to $\CircuitInput$, has a relativizing rewindable interactive proof with preprocessing (\cref{rem:rip-preprocessing}) with the following properties, where $\SoftO{\cdot}$ hides factors polylogarithmic in $\EncodedWitnessLength$ and running times count field operations.

  \begin{itemize}[noitemsep]
    \item \defn{Perfect completeness}.
    \item \defn{Soundness error}: at most $2^{\MathScriptStyle -\SoundnessParameter}$.
    \item \defn{Round complexity}: $\BigO{\SoundnessParameter\log\EncodedWitnessLength}$.
    \item \defn{Communication complexity}: $\SoftO{\SoundnessParameter/\eps}$ field elements.
    \item \defn{Verifier oracle query complexity}: $\BigO{\SoundnessParameter\min\{\OracleGateCount,1/\eps\}}$.
    \item \defn{Verifier preprocessing time}: $\SoftO{\EncodedWitnessLength}$.
    \item \defn{Verifier online time}: $\SoftO{\SoundnessParameter(\InstanceSize+\OutputSize+1/\eps)}$.
    \item \defn{Prover time}: $\BigO{\SoundnessParameter\EncodedWitnessLength}+\SoftO{\SoundnessParameter/\eps}$.
  \end{itemize}

\end{thm}

In particular, for constant $\SoundnessParameter$ and $1/\eps\le\log^{\MathScriptStyle O(1)}\EncodedWitnessLength$, the prover runs in time $\BigO{\EncodedWitnessLength}$, linear in the size of the computation, and the verifier's online time is $\SoftO{\InstanceSize+\OutputSize}$; the number of oracle queries, $\BigO{\min\{\OracleGateCount,1/\eps\}}$, is sublinear in $\OracleGateCount$ whenever $\eps\OracleGateCount$ grows without bound.

The proof applies the transformation of \cref{rel-iop-to-rip} to the IOP of \cref{ewiop-to-reliop} and reduces the soundness error to $2^{-\SoundnessParameter}$ by sequential repetition, restoring the prover to its initial state before each run. We then remark on computing proof strings on demand (\cref{rem:complexity-preserving}).

\begin{proof}
  Apply \cref{rel-iop-to-rip} to the IOP of \cref{ewiop-to-reliop}. The result is a relativizing rewindable interactive proof with preprocessing with soundness error at most $1/3$, $\BigO{\log\EncodedWitnessLength}$ rounds, $\BigO{\min\{\OracleGateCount,1/\eps\}}$ verifier oracle queries, communication $\SoftO{1/\eps}$, verifier online time $\SoftO{\InstanceSize+\OutputSize+1/\eps}$, verifier preprocessing time $\SoftO{\EncodedWitnessLength}$, and prover time $\BigO{\EncodedWitnessLength}+\SoftO{1/\eps}$. The verifier runs it $\RepetitionCount\coloneq\lceil\SoundnessParameter/\log_{2}3\rceil$ times sequentially with fresh randomness, restoring the prover to its initial state before each run, and accepts if and only if every run accepts. Perfect completeness is preserved. For soundness, fix a false instance and a deterministic cheating prover $\Malicious{\pro}$. Since each run queries $\Malicious{\pro}$ only on histories of that run, the runs are independent executions of the protocol against the same prover, each accepted with probability at most $1/3$, so all are accepted with probability at most $3^{\MathScriptStyle -\RepetitionCount}\le2^{\MathScriptStyle -\SoundnessParameter}$. The round complexity, communication, oracle queries, online time, and prover time are $\RepetitionCount$ times those of one run, and the preprocessing is done once, which gives the stated parameters.
\end{proof}

\begin{remark}[Computing proof strings on demand]\label{rem:complexity-preserving}\label{rem:linear-time-fft}
  Bitansky and Chiesa \cite{bitansky2012succinct} observed that the provers of a multi-prover interactive proof \cite{mip1988} (\cref{app:mip}) need not store a proof: each only ever returns single symbols of proof strings, so it can compute each symbol when it is queried. Holmgren and Rothblum \cite{holmgren2018delegating} sharpened this to provers with time $\SoftO{T}$ and space $S+o(S)$ for RAM computations. The same ability to compute symbols on demand is available to a rewindable prover, which also only returns single symbols: its state at the posting of a string already determines the answer at every position, and restoring that state before each query keeps the answer from depending on anything later, so the honest prover can fix a proof string without computing or storing it in full. This is an advantage over hash-based SNARGs: in a Merkle-based compilation of an IOP into a SNARG such as Flock's \cite{bunz2026flock}, the prover must compute every proof string in full in order to hash it. Computing on demand can be used to save prover time or prover space:
  \begin{itemize}[noitemsep]
    \item \defn{Prover time}. Most known linear-time provers come from tensor and interleaved codes of linear-time encodable codes \cite{bootle2020linear,ronzewi2022proving,bootle2022linear,bootle2022automata,xie2022orion,golovnev2023brakedown,brehm2025blaze,bunz2026tensorswitch}, as does the prover of \cref{thm:robust-rip-parameters}, which specifically uses TensorSwitch \cite{bunz2026tensorswitch} (\cref{lem:ild-encoding}). The concretely fastest IOPs, including Flock's, which our implementation builds on (\cref{sec:implementation}), encode with Reed--Solomon codes instead, whose encoding relies on the fast Fourier transform (FFT) and makes their provers quasi-linear; the linear-time instantiation is at present a theoretical result. Computing on demand avoids part of the FFT: our compilation of Flock's IOP applies only the first layers of the FFT to every row when it posts a proof string and finishes the remaining layers of a row only when the verifier asks for it. Together with the removal of the hash function and the smaller security parameter, this accounts for the speedup of our prover over Flock's SNARG prover in \cref{sec:impl-blake3}.
    \item \defn{Prover space}. \Cref{thm:robust-rip-parameters} bounds the prover's time but not its space, and as described the honest prover stores the encoding $\EWEncoding(\CandidateTable)$ of its witness, which takes space $\SoftO{\EncodedWitnessLength}$. Suppose that $\OracleCircuit$ computes the $T$-step execution of an oracle machine that uses space $S$, so that $\EncodedWitnessLength=\SoftO{T}$ and the prover can regenerate $\CandidateTable$ in space $\BigO{S}$ by running the machine again. Then, for the Reed--Solomon instantiation, each message of the prover can be computed in one pass over the regenerated $\CandidateTable$, as in the streaming prover of Cormode, Mitzenmacher, and Thaler \cite{cormode2012practical}, so the prover runs in space $\SoftO{S+1/\eps}$ and in time $\SoftO{\SoundnessParameter T}$, at the price of running the machine, and hence querying $\Oracle$, once per round. The verifier is unchanged. We do not work out the details here.
  \end{itemize}
\end{remark}

\section{Implementation and Experiments}\label{sec:implementation}

As a proof of concept, we implement our protocol and evaluate it on an
MNIST task inspired by the AI debate experiment of Irving, Christiano,
and Amodei \cite{irving2018ai}. Briefly, in their experiment, there
are two provers, or agents, and a judge which is trained to
classify an MNIST image from a few nonzero pixels. A random MNIST image
is shown to the agents but not to the judge. One agent claims the
correct label and the other agent claims an incorrect label, e.g., the
image contains a 5, and one claims 5 and the other claims 6. They take
turns revealing a nonzero pixel in the image to the judge until
four or six pixels are revealed. Then the judge chooses which agent
wins based on which of their labels has the higher logit. Their experiment
studies how often an honest agent wins against a dishonest one; we
instead measure the cost of verifying a robust computation that uses
their judge as its oracle.

Following their experiment, we train a sparse MNIST classifier using
the same architecture; it correctly classifies $59.4\%$ of random
six-pixel views. We use its predictions to define an oracle:
given a sparse view and a candidate digit $y$, the oracle outputs $1$
if the classifier predicts $y$ and $0$ otherwise. Our circuits query
this oracle on many sparse views of an image and output whether a
majority of the views vote for the candidate digit. We vary the
number of sparse views to study how the cost of verification scales
with the size of the circuit. We also compile an existing IOP for a
standard, oracle-free computation, Flock's IOP for BLAKE3
\cite{bunz2026flock}, into an RIP and compare it with
Flock's own SNARG (\cref{sec:impl-blake3}).

\subsection{Implementation}\label{sec:impl-overview}

Our implementation is open source\footnote{\url{https://github.com/ZiyiGuan/oversight-rip}} and written in Rust. It follows the construction of
\cref{sec:robust}, building the encoded-witness IOP of
\cref{sec:ewiop} from components of Binius \cite{diamond2023binius},
Flock \cite{bunz2026flock}, and Ligero \cite{ames2017ligero}, and
compiling it into an RIP as in \cref{rel-iop-to-rip}: the prover keeps
each proof string it computes instead of sending it, and the verifier
reads a symbol of a proof string by restoring the prover to its state
at the moment that string was fixed and asking for that symbol, so that
no answer can depend on anything the verifier sent later. The proof
strings are opened in the layout of Ligero \cite{ames2017ligero}: the
encoded witness is arranged in rows of about $\sqrt{\TranscriptLength}$
entries and the verifier reads $64$ whole columns, which gives a
proximity error below $2^{-42}$. Together with $\SampleCount$ sampled
oracle gates, the smallest number with $(1-\eps)^{\SampleCount}\le
2^{-41}$ for the circuit's robustness parameter $\eps$, a \emph{single}
execution of the protocol has soundness error at most $2^{-40.53}$ at
every circuit size, against computationally unbounded provers; there
are no repetitions.

The majority circuit sums the oracle answers in a tree whose levels
consist of identical blocks, so the verifier's work for a level is that
of one block and this circuit family needs no preprocessing; for a general
circuit, the preprocessing of \cref{rem:rip-preprocessing} would take
its place. The
classifier's answers are computed in advance, so both the protocol and
direct evaluation of the circuit query the oracle by table lookup. All
timings are from AWS Graviton4 instances, one single-threaded process
per core, with transparent huge pages enabled. Absolute times are
specific to that core; every ratio below is between measurements on
the same core type.

\subsection{MNIST experiments}

Concretely, call an MNIST image with six nonzero pixels revealed a
view. For each circuit, we fix an MNIST image together with $q_C$
randomly sampled views of the image; the image and views are
hardwired into the description of the circuit. On input a candidate
digit $y$, the circuit queries the oracle on each of its $q_C$ views
and outputs $1$ if more than half of the views vote for $y$. We can
compute the robustness parameters for these circuits exactly: for each
digit $y$, we compute the minimum number of oracle answers that would
need to be changed in order to flip the output of the circuit on
$y$. Let $m$ be the minimum of this quantity over all ten digits. Then
changing at most $m-1$ oracle answers cannot change the output of the
circuit on any input, so the circuit is
$(m-1)/\OracleGateCount$-robust with respect to every input. 

To choose instances to run our circuit on, we scan the MNIST test set
in order. For each image, we sample $2^{8}$ random views and construct
the corresponding circuit. We select the first 50 images for which
this circuit accepts the correct digit and has robustness parameter at
least $1/8$.  We use these same 50 images throughout the experiment,
and consider
\[
\OracleGateCount \in
\{2^{8},2^{10},2^{12},\ldots,2^{26}\},
\]
sampling additional views of each image as $\OracleGateCount$ increases
(one image has fewer than $2^{24}$ distinct six-pixel views and three have
fewer than $2^{26}$; those images are omitted at those sizes). For each
image and circuit size, we recompute the robustness parameter and run
the complete protocol once, at soundness error
$2^{-40}$. The median robustness parameter remains between $0.30$ and
$0.32$ at every circuit size.

All runs with the honest
provers accepted, as expected from perfect completeness. \cref{tab:mnist-cost} reports detailed costs, and \cref{fig:mnist-plots} shows how the verifier scales relative to direct evaluation. All values are medians over the images.

Since the robustness parameter
remains nearly constant, the median number of oracle queries made by
the verifier stays between $74$ and $80$ as $\OracleGateCount$ grows
from $2^{8}$ to $2^{26}$. At $2^{26}$ views, the verifier makes
$840{,}000\times$ fewer oracle queries than the circuit. The verifier's
online time also grows much more slowly than direct evaluation: it
grows from $0.45$\,ms to $1.9$\,ms, while direct evaluation grows
from $0.06\,\mu$s to $16$\,ms. Because direct evaluation costs only
about $0.23$\,ns per view, the two cross between $2^{22}$ and $2^{24}$
views, and at $2^{26}$ the verifier takes $0.12\times$ as long as direct
evaluation. This baseline is pessimistic for the protocol: both direct
evaluation and the protocol answer oracle queries by table lookup,
whereas in the intended setting each query is a model evaluation and
dominates everything else. With a costly oracle, direct evaluation pays
$q_C$ evaluations while the verifier pays his $74$ to $80$, so the
crossing would come at a far smaller circuit, and the prover's overhead
over direct evaluation would shrink toward the cost of its field
arithmetic alone; the oracle-query counts in \cref{tab:mnist-cost} are
the measure that transfers.

\begin{figure}[!htbp]
  \centering
  \begin{tikzpicture}
    \begin{loglogaxis}[
      width=0.48\linewidth, height=0.36\linewidth,
      xlabel={$\OracleGateCount$}, ylabel={time (ms)},
      log basis x=2, xtick={256,4096,65536,1048576,16777216},
      xticklabels={$2^{8}$,$2^{12}$,$2^{16}$,$2^{20}$,$2^{24}$},
      legend pos=north west, legend cell align=left, legend style={font=\small, draw=none, fill=none},
      tick label style={font=\small}, label style={font=\small},
      mark size=1.6pt]
      \addplot+[mark=*] coordinates {(256,0.446) (1024,0.495) (4096,0.522) (16384,0.643) (65536,0.689) (262144,0.8) (1048576,0.945) (4194304,1.11) (16777216,1.5) (67108864,1.88)};
      \addplot+[mark=square*, color=red] coordinates {(256,6.4e-05) (1024,0.000238) (4096,0.000936) (16384,0.00373) (65536,0.0149) (262144,0.0596) (1048576,0.238) (4194304,0.953) (16777216,3.83) (67108864,15.5)};
      \legend{verifier online, direct evaluation}
    \end{loglogaxis}
  \end{tikzpicture}
  \hfill
  \begin{tikzpicture}
    \begin{loglogaxis}[
      width=0.48\linewidth, height=0.36\linewidth,
      xlabel={$\OracleGateCount$}, ylabel={oracle queries},
      log basis x=2, xtick={256,4096,65536,1048576,16777216},
      xticklabels={$2^{8}$,$2^{12}$,$2^{16}$,$2^{20}$,$2^{24}$},
      log basis y=2, ytick={64,1024,16384,262144,4194304},
      yticklabels={$2^{6}$,$2^{10}$,$2^{14}$,$2^{18}$,$2^{22}$},
      legend pos=north west, legend cell align=left, legend style={font=\small, draw=none, fill=none},
      tick label style={font=\small}, label style={font=\small},
      mark size=1.6pt]
      \addplot+[mark=*] coordinates {(256,76) (1024,74) (4096,75) (16384,75) (65536,76) (262144,76) (1048576,76) (4194304,76) (16777216,78) (67108864,80)};
      \addplot+[mark=square*, color=red] coordinates {(256,256) (1024,1024) (4096,4096) (16384,16384) (65536,65536) (262144,262144) (1048576,1048576) (4194304,4194304) (16777216,16777216) (67108864,67108864)};
      \legend{verifier, direct evaluation}
    \end{loglogaxis}
  \end{tikzpicture}
  \caption{Verifier cost compared with direct circuit evaluation as the
number $q_C$ of oracle gates increases, for one execution of the
protocol at soundness error $2^{-40}$. Curves are medians over the
selected MNIST images (50, or 49 and 47 at the two largest sizes). Left: verifier online time compared with one
direct circuit evaluation. Right: verifier oracle queries compared with
the $q_C$ queries made by direct evaluation. Both axes are logarithmic.}
  \label{fig:mnist-plots}
\end{figure}

At small $q_C$, the verifier is slower than direct evaluation because protocol overhead dominates: the verifier performs
the IOP interaction and opening checks, whereas
direct evaluation requires only $q_C$ table lookups. As $q_C$ grows,
direct evaluation grows linearly, while the verifier's time grows by a
factor of $1.1$ to $1.3$ per fourfold increase in $q_C$, so their
ratio falls by a factor of $3.0$ to $3.7$ per step. The verifier
time grows because our implementation uses the
Ligero layout~\cite{ames2017ligero}, whose openings have size about
$\sqrt{N}$, rather than the polylogarithmic cost in \cref{thm:robust-rip-parameters}.

The prover cost settles at a constant multiple of direct evaluation from
$\OracleGateCount=2^{20}$ on: the prover, which generates the witness
and makes all $q_C$ oracle lookups, takes about $60\times$ direct
evaluation ($0.9$\,s at $2^{26}$ views); at the small sizes it is
dominated by fixed costs.

\begin{table}[!htbp]
  \centering
  \small
  \begin{tabular}{@{}rrrrr@{}}
    \toprule
    & & & \multicolumn{2}{c}{multiple of direct eval.} \\
    \cmidrule(lr){4-5}
    $\OracleGateCount$ & $\ver$ oracle queries & direct eval. & $\ver$ online & $\pro$ \\
    \midrule
    $2^{8}$ & 76 & 0.064\,$\mu$s & 6{,}976$\times$ & 4{,}809$\times$ \\
    $2^{10}$ & 74 & 0.238\,$\mu$s & 2{,}081$\times$ & 1{,}635$\times$ \\
    $2^{12}$ & 75 & 0.936\,$\mu$s & 558$\times$ & 567$\times$ \\
    $2^{14}$ & 75 & 3.73\,$\mu$s & 172$\times$ & 237$\times$ \\
    $2^{16}$ & 76 & 14.9\,$\mu$s & 46.2$\times$ & 117$\times$ \\
    $2^{18}$ & 76 & 59.6\,$\mu$s & 13.4$\times$ & 79.5$\times$ \\
    $2^{20}$ & 76 & 238.3\,$\mu$s & 3.97$\times$ & 64.8$\times$ \\
    $2^{22}$ & 76 & 953.1\,$\mu$s & 1.17$\times$ & 60.9$\times$ \\
    $2^{24}$ & 78 & 3.83\,ms & 0.39$\times$ & 59.2$\times$ \\
    $2^{26}$ & 80 & 15.51\,ms & 0.12$\times$ & 58.8$\times$ \\
    \bottomrule
  \end{tabular}
  \caption{Cost of the protocol on the MNIST circuits, one execution at soundness error $2^{-40}$ (Graviton4, one core): medians over the 50 images, or over those with enough distinct views at the two largest sizes. ``Direct eval.'' is the time to evaluate the circuit directly, one table lookup per view and the majority. Verifier online time and prover time are reported as multiples of it; there is no preprocessing. $\pro$ includes witness generation.}
  \label{tab:mnist-cost}
\end{table}

\subsection{Comparison with a SNARG}\label{sec:impl-blake3}

In \cref{sec:intro} we argued that rewinding brings efficiency
gains over a SNARG. The relativizing protocols of
Chen, Kalai, and Xi \cite{chen2026avoid} offer no comparison, since they
have no implementation and their prover is only polynomial-time, so we
compare prover time against the best known concretely efficient
proof system for ordinary, oracle-free computations,
Flock \cite{bunz2026flock}, whose IOP proves batches of BLAKE3
compressions and whose SNARG, a succinct non-interactive argument,
compiles it with Merkle commitments. A SNARG is how an IOP is made
practical in the applications of proof systems to date, where the
verifier has no control over the prover's machine and so cannot rewind
it: the IOP's proof strings are committed to with a hash function and
the interaction is removed. Rewinding is another way to make the same
IOP practical, so comparing our compilation with Flock's SNARG measures
what each way costs. We
compiled Flock's IOP into an RIP, keeping its proof strings, queries,
and checks: the prover runs
Flock's prover with the Merkle commitments removed, and the verifier
rewinds it to its state at each proof string's posting before every row
query, as in the compilation of \cref{rel-iop-to-rip}, so that its
answers depend on the posting alone; the verifier runs Flock's verifier
on those rows. A false statement is then
accepted with at most the IOP's own error (\cref{rel-iop-to-rip}).

The RIP gains in three ways over the SNARG. It needs no hash
function, since rewinding replaces the Merkle commitment. Nor does it
need to encode every proof string in full, since nothing is hashed: the
prover applies only the first layers of the FFT when it posts a string
and finishes a row only when it is queried (\cref{rem:linear-time-fft}). And
it needs far fewer bits of security: a SNARG prover can retry offline
until the hash-derived challenges are favorable, so Flock's default
configuration targets a $128$-bit per-round work factor, whereas a
cheating prover in the RIP gets one attempt per
execution, which the verifier observes, so $40$ bits suffice. We therefore run
Flock's IOP at its $40$-bit interactive parameter setting and, since a
smaller field suffices as well, over a $64$-bit base field with a
$128$-bit extension for the challenges, which Flock's release does not
offer and which halves the field arithmetic. The error is $2^{-42.4}$
to $2^{-42.9}$ at these sizes, assuming that the IOP's query and
algebraic error terms compose statistically, as Flock's parameter code
assumes.

\cref{tab:blake3} reports prover times at
$2^{22}$ to $2^{34}$ witness bits ($2^{8}$ to $2^{20}$ compressions) on
one Graviton4 core, medians of three executions. Our prover is $1.8$
to $2.1\times$ faster than Flock's SNARG prover at every size.

\begin{table}[!htbp]
  \centering
  \small
  \begin{tabular}{@{}rrrr@{}}
    \toprule
    witness bits & Flock's SNARG prover & our prover & speedup \\
    \midrule
    $2^{22}$ & 89.9\,ms & 46.7\,ms & 1.92$\times$ \\
    $2^{24}$ & 101.6\,ms & 54.7\,ms & 1.86$\times$ \\
    $2^{26}$ & 153.9\,ms & 86.9\,ms & 1.77$\times$ \\
    $2^{28}$ & 480.1\,ms & 223.7\,ms & 2.15$\times$ \\
    $2^{30}$ & 1.23\,s & 694.2\,ms & 1.77$\times$ \\
    $2^{32}$ & 4.81\,s & 2.66\,s & 1.81$\times$ \\
    $2^{34}$ & 19.8\,s & 10.6\,s & 1.86$\times$ \\
    \bottomrule
  \end{tabular}
  \caption{Prover time for Flock's IOP for BLAKE3 compiled as Flock's
SNARG (default configuration, Merkle commitments, sound against bounded
provers) and as our protocol over a $64$-bit base field at
the $40$-bit interactive parameters, with the prover rewound
before each query (sound against unbounded provers with the IOP's own
error, $2^{-42.4}$ to $2^{-42.9}$). The speedup is Flock's SNARG prover's time divided
by ours. Medians of three executions on one Graviton4 core, including
witness generation.}
  \label{tab:blake3}
\end{table}

\subsection{Limitations}

Our experiments are intended as a proof of concept rather than as a
realistic scalable-oversight benchmark. The circuits are simple
majority circuits over random sparse views, and the 50 images are
selected so that their $\OracleGateCount=2^{8}$ circuits have robustness parameter at
least $1/8$; thus, the experiments do not show that arbitrary MNIST
instances or more complex oracle computations satisfy our robustness
assumption. Moreover, unlike in the MNIST debate experiment, the
sparse views are sampled randomly instead of chosen adversarially (by
debating agents). The majority is also a statistical quantity: for a
majority over hardwired views, the verifier could estimate the output
himself by sampling about as many views as he queries in the protocol,
with no prover at all. The protocol is needed for circuits that process
the oracle answers further or choose later queries from earlier answers;
our circuits do neither.

Because the image and views are hardwired into the circuit, each image
and circuit size is its own instance; the verifier needs no
preprocessing here only because the majority circuit's levels consist of
identical blocks, and a general circuit would need the preprocessing of
\cref{rem:rip-preprocessing}, with its own amortization question. The
verifier's time grows with the square root of the
trace, from the Ligero layout of the opening, rather than
polylogarithmically as in \cref{thm:robust-rip-parameters}. Even though
the verifier becomes cheaper than direct evaluation on the largest
circuits, the prover cost remains a constant factor over direct
evaluation, about $60\times$ at the largest sizes,
where direct evaluation is $q_C$ table lookups and the prover performs
field arithmetic over the entire encoded witness; on the BLAKE3
instances, where there is no such gap to begin with, our prover
is cheaper than the SNARG's. The protocol also treats saving and restoring the
prover's state as free; for a real model each restore copies its
state, and in both compilations the prover is restored before every
query.

\section*{AI Disclosure}
\addcontentsline{toc}{section}{AI Disclosure}
We used generative AI tools to write large portions of the code for our implementation and experiments, draft and edit sections of this paper, proofread the paper to catch potential bugs and suggest writing improvements, do a literature review, and help with numerical calculations. All AI-generated text and code were manually reviewed and verified by the authors. For example, we tested AI-generated code and checked consistency against the theoretical protocol, and we did a manual literature survey. The theoretical model and construction were developed solely by the authors. We take full responsibility for the content of this work.  

\section*{Acknowledgments}
\addcontentsline{toc}{section}{Acknowledgments}
This work is supported by the U.S.\ National Science Foundation under award No.~2534400, and by the UK AISI alignment project.  Ron Rothblum was partially supported by the Advanced Research and Invention Agency (ARIA). We would like to also thank Benedikt B\"{u}nz, Alessandro Chiesa, Giacomo Fenzi, Christian Knabenhans, William Wang, and Andrew Zitek for helpful discussions on the implementations of the protocols, and Ryan Williams and Nathan Sheffield for a helpful discussion on properties of proof systems specific to scalable oversight. 

\printbibliography[heading=bibintoc]

\appendix

\section{Multi-prover interactive proofs}\label{app:mip}
Rewindable interactive proofs are closely related to multi-prover interactive proofs (MIPs) \cite{mip1988}: a restored snapshot gives the verifier an interaction that the prover cannot connect to any other, much as a second non-communicating prover would. MIPs are in turn equivalent in power to probabilistically checkable proofs (PCPs) \cite{feige1997making,arora1998probabilistic,arora1998proof}, since two non-communicating provers can simulate a proof string, at a loss in soundness, and a proof string can hold the provers' strategies \cite{fortnow1994power}; we therefore present only the equivalence of MIPs and RIPs. This appendix recalls the definition of MIPs and shows that the two models have the same power (\cref{sec:rip-vs-mip}): every MIP is an RIP with the same soundness error, and every RIP whose verifier makes $\RIPQueryCount$ queries to the prover is a two-prover MIP with soundness error $1-(1-\RIPSoundnessError)/\RIPQueryCount$.

A multi-prover interactive proof lets several powerful provers try to convince a weak verifier that an instance has a valid witness.

\begin{definition}[Multi-prover interactive proof system]\label{defn:mip}

  A $\ProverCount$-prover interactive proof for a relation $\Relation$ is an interactive protocol between a probabilistic polynomial-time verifier $\ver$ and $\ProverCount$ computationally unbounded provers $\pro_{\MathScriptStyle 1},\ldots,\pro_{\MathScriptStyle \ProverCount}$. The parties receive a common input $\Instance$, and the honest provers additionally receive a witness $\Witness$ with $(\Instance,\Witness)\in\Relation$. 

  The provers may agree on a strategy beforehand but cannot communicate during the protocol: each prover's messages depend only on $\Instance$, its agreed strategy, and its own interaction with the verifier. At the end of the interaction, the verifier accepts or rejects.

  The protocol satisfies the completeness and soundness properties defined below.

\end{definition}

We write $\langle\pro_{\MathScriptStyle 1}(\Witness),\ldots,\pro_{\MathScriptStyle \ProverCount}(\Witness),\ver\rangle(\Instance)=1$ when the verifier accepts after interacting with the provers on common input $\Instance$. The probabilities below are over the verifier's private random bits (we assume without loss of generality that the provers are deterministic). 

\begin{definition}[Completeness]\label{definition:mip-completeness}

  An MIP for $\Relation$ has \defn{perfect completeness} if, for every $(\Instance,\Witness)\in\Relation$, the honest provers satisfy
  \[
    \probability{
      \langle\pro_{\MathScriptStyle 1}(\Witness),\ldots,\pro_{\MathScriptStyle \ProverCount}(\Witness),\ver\rangle(\Instance)=1
    }=1.
  \]

\end{definition}

\begin{definition}[Soundness]\label{definition:mip-soundness}

  An MIP for $\Relation$ has \defn{soundness error} $\MIPSoundnessError$ if, for every $\Instance\notin\Language$ and every tuple of possibly computationally unbounded cheating provers $\Malicious{\pro}_{\MathScriptStyle 1},\ldots,\Malicious{\pro}_{\MathScriptStyle \ProverCount}$,
  \[
    \probability{
      \langle\Malicious{\pro}_{\MathScriptStyle 1},\ldots,
      \Malicious{\pro}_{\MathScriptStyle \ProverCount},\ver\rangle(\Instance)=1
    }\le\MIPSoundnessError(|\Instance|).
  \]

\end{definition}

\begin{definition}[Relativizing MIP]\label{defn:relativized-mip}

  Fix an oracle $\Oracle$ and a relativized relation $\OracleRelation$ as in \cref{defn:relativized-relation}. A \defn{relativizing MIP} for $\OracleRelation$ is an MIP for $\OracleRelation$ in which the verifier and all provers, including cheating provers, have black-box access to the same oracle $\Oracle$. The protocol \defn{relativizes} if its completeness, soundness, and efficiency guarantees hold for $\OracleRelation$ for every oracle $\Oracle$.

\end{definition}

\begin{remark}[Preprocessing]\label{rem:mip-preprocessing}
  As in \cref{rem:iop-preprocessing}, an MIP \defn{with preprocessing} for instances of the form $\Instance=(\InstancePre,\InstancePost)$ has a verifier that is a pair $\ver=(\VerifierPre,\VerifierPost)$. First $\VerifierPre$ receives $\InstancePre$ and outputs a state $\VerifierState$. Then $\VerifierPost$ receives $\VerifierState$ and $\InstancePost$ and interacts with the provers. Completeness and soundness are as in \cref{definition:mip-completeness,definition:mip-soundness}.
\end{remark}

Efficiency parameters are as for RIPs (\cref{sec:rip-definition}), with parallel exchanges with different provers counting as one round and the prover running time taken as the sum over the honest provers.

\subsection{Relation to RIPs}\label{sec:rip-vs-mip}
Rewindable interactive proofs are as powerful as MIPs: every MIP is a rewindable interactive proof with the same soundness error, and two provers can simulate a rewindable prover at a loss in the soundness error.

\begin{lemma}[MIP to RIP]\label{lem:mip-to-rewindable}

  Every $\ProverCount$-prover interactive proof for $\Relation$ with perfect completeness and soundness error $\MIPSoundnessError$ yields a rewindable interactive proof for $\Relation$ with perfect completeness and soundness error $\RIPSoundnessError=\MIPSoundnessError$, the same verifier complexity up to tagging each message with a prover index, and the same prover running time. If the MIP has preprocessing (\cref{rem:mip-preprocessing}), so does the RIP (\cref{rem:rip-preprocessing}), with the same verifier preprocessing. The same holds for relativizing protocols.

\end{lemma}

\begin{proof}

  Let $(\pro_{\MathScriptStyle 1},\ldots,\pro_{\MathScriptStyle \ProverCount},\ver)$ be the MIP. The new verifier runs $\ver$, and whenever $\ver$ has sent messages $\VerifierMessage_{\MathScriptStyle 1},\ldots,\VerifierMessage_{\MathScriptStyle j}$ to prover $i$, it obtains that prover's reply by querying the rewindable prover on the history $((i,\VerifierMessage_{\MathScriptStyle 1}),\ldots,(i,\VerifierMessage_{\MathScriptStyle j}))$. The honest rewindable prover answers a history tagged with $i$ as $\pro_{\MathScriptStyle i}$ would, so completeness is preserved. For soundness, given any function $\Malicious{\pro}$, define cheating MIP provers by $\Malicious{\pro}_{\MathScriptStyle i}(\VerifierMessage_{\MathScriptStyle 1},\ldots,\VerifierMessage_{\MathScriptStyle j})\coloneq\Malicious{\pro}((i,\VerifierMessage_{\MathScriptStyle 1}),\ldots,(i,\VerifierMessage_{\MathScriptStyle j}))$. Each $\Malicious{\pro}_{\MathScriptStyle i}$ depends only on its own history, so these are valid non-communicating MIP provers, and the new verifier's view against $\Malicious{\pro}$ is exactly that of $\ver$ against $\Malicious{\pro}_{\MathScriptStyle 1},\ldots,\Malicious{\pro}_{\MathScriptStyle \ProverCount}$. Hence the acceptance probability is at most $\MIPSoundnessError$, so $\RIPSoundnessError\le\MIPSoundnessError$. The new verifier preprocesses $\InstancePre$ exactly as $\ver$ does, and the same argument applies verbatim when all parties have access to an oracle $\Oracle$.

\end{proof}

\begin{lemma}[RIP to MIP]\label{lem:rewindable-to-mip}

  Every rewindable interactive proof for $\Relation$ with perfect completeness and soundness error $\RIPSoundnessError$ in which the verifier makes at most $\RIPQueryCount$ queries to the prover yields a two-prover interactive proof for $\Relation$ with perfect completeness and soundness error $\MIPSoundnessError\le1-(1-\RIPSoundnessError)/\RIPQueryCount$, with one additional round, in which the first prover answers the verifier's queries and the second prover answers one query. The same holds for relativizing protocols and for protocols with preprocessing.

\end{lemma}

\begin{proof}

  The verifier runs the RIP, sending each query to the first prover, and at the end sends one of the queries, chosen uniformly at random, to the second prover and rejects if the two answers differ. With honest provers the answers always agree. For soundness, the second prover's answers define a single next-message function, hence a cheating RIP prover $\Malicious{\pro}$. Let $C$ be the event that the first prover's answers agree with $\Malicious{\pro}$ on every queried history. On $C$ the simulated RIP verifier's view is its view against $\Malicious{\pro}$, so $\probability{\text{accept}\wedge C}\le\min\{\RIPSoundnessError,\probability{C}\}$. Otherwise some answer disagrees, and the second prover's query, chosen independently of the transcript, lands on it with probability at least $1/\RIPQueryCount$, so $\probability{\text{accept}\wedge\neg C}\le(1-1/\RIPQueryCount)\probability{\neg C}$. Writing $b\coloneq\probability{\neg C}$, the acceptance probability is at most $\min\{\RIPSoundnessError,1-b\}+b(1-1/\RIPQueryCount)\le1-(1-\RIPSoundnessError)/\RIPQueryCount$, with the maximum at $b=1-\RIPSoundnessError$.

\end{proof}

\end{document}